# Negative Ontology of True Target for Machine Learning: Towards Recognition, Evaluation and Learning under Democratic Supervision

Yongquan Yang[1*]

[1]Institute of Sciences for AI, Chengdu, Sichuan, China

[*]Corresponding author (Email: remy_yang@foxmail.com or yongquan.yang@sciences4ai.com; ORCiD: 0000-0002-3965-4816)

This article philosophically examines how a shift in the assumed ontology of the true target (TT) can lead to a new paradigm for machine learning (ML)-based predictive modelling. By systematically analysing the existence assumption of the TT underlying mainstream ML paradigms, we adopt a negative ontology perspective, explicitly positing that the TT does not objectively exist in the real world as a universally accessible object. On this basis, we define Democratic Supervision as an alternative supervisory principle for ML, in which no single source is assumed to possess an objectively privileged target. We further introduce Multiple Inaccurate True Targets (MIATTs) as an instance-level realization of Democratic Supervision. Building upon MIATTs, we establish the logic-driven generation and assessment for MIATTs construction (recognition with MIATTs), formulate logical assessment formula for evaluation with MIATTs, and develop undefinable true target learning for learning with MIATTs. These components are integrated to formulate the Recognition, Evaluation, Learning with MIATTs (REL-MIATTs) framework. We further characterize REL-MIATTs from the perspective of Cognitive Machine Learning: a single cycle provides an elementary cognitive learning unit, MIATTs introduce complementary supervisory perspectives, and iterative cycles support the evolution of learning through accumulated experience, feedback, and state updating. Extending this process to human–AI interaction provides a basis for human–AI co-cognitive development and, potentially, human–AI co-evolution. Using a synthetic controlled environment, we examine how the iterative recognition, evaluation, and learning processes of REL-MIATTs support human–AI co-evolution and adaptive knowledge discovery, providing a controlled reference for potential real-world applications. A real-world application further demonstrates the potential of the framework for supporting individual education and professional development, providing empirical evidence for the feasibility of Democratic Supervision, REL-MIATTs, and its broader implications for continuous human–AI co-evolution.

## Introduction

The true target (TT), which is a computationally equivalent transformation of the ground-truth, serves as a fundamental concept in the formulation and deployment of ML paradigms [1]. Assumptions regarding the TT are therefore crucial, as they implicitly define what is being learned, how supervision is interpreted, and how models are expected to approximate and generalize the underlying reality. (In this article, the term TT is used instead of ground-truth to facilitate practical discussion. The transformation from the ground-truth to a computational TT and its reverse are both essential in practice. Semantically, the two are equivalent.)

Despite their central role, TT assumptions are often taken for granted in mainstream evaluation and learning paradigms of ML, where the objective existence of a well-defined TT is treated as a default premise. However, growing evidence from evaluation with inaccurate true targets and noisy-label learning and multiple-annotator settings suggests that, in many real-world tasks, the TT may be ambiguous, subjective, or even inherently undefinable [2–8].

These observations expose a tension between the traditional existence assumption of TT and the practical realities of data acquisition and annotation. This tension motivates a deeper philosophical examination of the existence assumptions about TT for ML paradigm design [9].

The mainstream existence assumption of TT across major evaluation and learning paradigms is that the TT objectively exists in the real world. Evaluation and learning paradigms under inaccurate supervision are more likely to raise questions about the appropriateness of this existence assumption. A clear emerging trend is that a substantial body of work increasingly challenges the clarity, definability, and uniqueness of TT [3, 7, 8, 10–36]. Although the scepticism expressed in these studies typically manifests as the acknowledgment that the TT may not exist, few works directly reject the objective existence of the TT, regardless of its clarity, definability, or uniqueness (see Section 2).

In this article, we argue that claiming/acknowledging the TT may not exist does not amount to a direct rejection of its objective existence. A true rejection requires a stronger claim/acknowledgement. Accordingly, we adopt a more radical stance by explicitly positing the Negative Ontology of TT for ML: The TT does not objectively exist in the real world as a universally accessible object [9, 13]. Here, the phrase "a more radical stance" indicates a possible line of reasoning: even if a TT physically exists (e.g., in conventional supervised learning setting), one can still deliberately adopt the non-existence assumption of TT. This can be understood as a situation, in which a practitioner lacks any prior specification or understanding of what the TT is while exploring ML-based predictive modelling. Adopting such a stance of the non-existence assumption of TT may open up new perspectives and insights. For example, from the first principles, it reframes the question of how the ML-based predictive modelling should proceed when the practitioner initially has no knowledge of the TT.

The explicitly posited non-existence assumption of TT originates from our previous works on uncovering inherently ambiguous TT through ML in the medical field [9–14]. While carrying out these studies, we gradually realized that assuming the objective existence of TT was fundamentally inadequate for tasks characterized by the intrinsic indefinability of TT. A more appropriate assumption, we found, is that TT does not objectively exist in the real world as a universally accessible object. These works thus served as the conceptual foundation that eventually led us to recognize the central importance of the assumption concerning the existence of TT for ML. Furthermore, insights from our philosophical analysis of TT assumptions across major ML paradigms, together with the subsequent implications and comparisons between the non-existence and existence assumptions, confirm both the novelty and necessity of the non-existence assumption proposed in this article.

Under the non-existence assumption of TT, we define Democratic Supervision for ML. The term "Democratic Supervision" has been discussed in contexts such as education [37–40] and professional development [41]. In these two fields, Democratic Supervision is defined as a collaborative and participatory approach in which authority is shared and stakeholders jointly engage in decision-making and reflective practice to support collective and individual growth. Here, we introduce the "Democratic Supervision" term into ML and provide a formal definition grounded in the non-existence assumption of TT. Democratic Supervision for ML is defined as a supervision paradigm that, under the non-existence of TT, constructs supervisory signals through the aggregation, negotiation, or coexistence of diverse contributors' perspectives, rather than relying on a unique authoritative ground truth.

This perspective generalizes the conventional setting and opens a broader space for developing evaluation and learning paradigms under democratic/pluralistic supervision and leads to substantially different perspectives and insights on supervision and ML paradigms. Under the conventional existence assumption of TT, ML paradigms rely on domain expert–centric supervision, where paradigm design is largely constrained by the quality of expert-annotated data. In contrast, under the non-existence assumption of TT, Democratic Supervision reconceptualizes supervision as a participatory process involving domain experts, ML practitioners, and potentially non-experts, thereby enabling more flexible and inclusive data construction. This perspective generalizes the conventional settings (strong or weak supervision-based evaluation and learning) and opens a broader space for developing evaluation and learning paradigms under democratic or pluralistic supervision, extending ML research in this direction. It also introduces significant transformations and advantages at both the humanistic level and the technical level (see Section 3).

We present an operationalizing component, in which Multiple Inaccurate True Targets (MIATTs) is defined as the operation for Democratic Supervision at the instance level. MIATTs is formalized as a finite set of inaccurate true targets, each capturing only a subset of the semantic facts of an underlying (possibly undefinable) true target, such that no individual target is complete but their union collectively represents partial or full aspects of it. (See Section 4)

Based on MIATTs, we derive principles for Logic-Driven Generation and Assessment (LDGA) for MIATTs construction, which is formulated as a process of recognition with MIATTs under the absence of an objectively accessible TT[42]. The resulting MIATTs provide multiple supervisory perspectives for evaluation and learning. We further derive principles for a Logical Assessment Formulation (LAF) for evaluation with MIATTs [13] and for Undefinable True Target Learning (UTTL) for learning with MIATTs [9]. These components provide the foundation for establishing the recognition, evaluation and learning with MIATTs (REL-MIATTs) framework for ML-based predictive modelling [43]. The development of REL-MIATTs also connects the proposed framework to Cognitive Machine Learning (CML) [44]. While CML primarily starts from cognitive mechanisms, REL-MIATTs approaches cognitive learning from the Negative Ontology of TT assumption. From this perspective, a single REL-MIATTs cycle can be viewed as an elementary cognitive learning unit, while MIATTs and iterative REL-MIATTs exhibit characteristics associated with the emergence, complementarity, and evolution of learning, respectively. Extending iterative REL-MIATTs to human–AI interaction further provides a basis for human–AI co-cognitive development and, potentially, human–AI co-evolution. (see Section 5)

Beyond extending ML from target-oriented prediction toward adaptive exploration and refinement of possible representations/hypotheses of underlying structural realities, REL-MIATTs and its iterative extension provide a basis for re-examining the theoretical foundations and practical implications of ML from complementary perspectives. Accordingly, we investigate both the broader theoretical implications and potential applications of REL-MIATTs, for human-centred adaptive knowledge discovery with potential applications including explanation discovery, causal exploration, scientific hypothesis generation, and human–AI collaborative knowledge evolution (see Sections 6 and 7).

The comparison between REL-MIATTs and CML suggests that REL-MIATTs provides a potential basis for Human–AI Co-Evolution, while the theoretical implications and emerging applications suggest its potential for human-centred Adaptive Knowledge Discovery. Accordingly, we employ a synthetic controlled environment (SCE) to examine how the iterative recognition, evaluation, and learning processes of REL-MIATTs may enable Human–AI Co-Evolution and Adaptive Knowledge Discovery. Notably, with successive mutual Human-AI Co-Evolution, the REL-MIATTs system eventually 'jumps' to be close to the explanatory structure that does not objectively exist in the SCE as a universally accessible object. These findings provide controlled evidence for the behavioural and evolutionary principles of REL-MIATTs and a reference for understanding its potential application for personalized education, professional development, and lifelong learning in real-world scenarios (see Section 8).

The principles of LDGA motivate the development of concrete algorithms for MIATTs generation, assessment, and task-difficulty estimation (MIATTs construction / recognition with MIATTs). This is a fundamental prerequisite for realizing REL-MIATTs, as MIATTs provide the core supervisory structures that substitute for inaccessible true targets under the Negative Ontology of TT assumption. Accordingly, we develop LDGA-based algorithms for MIATTs generation and assessment [42], together with an MIATTs-based algorithm for task-difficulty estimation, thereby transforming the key components of REL-MIATTs into executable procedures. Beyond the MIATTs construction, the practical realization of REL-MIATTs relies on two complementary mechanisms for real-world machine learning tasks: LAF-based evaluation with and UTTL-based learning with MIATTs. We therefore develop LAF-grounded evaluation strategies and UTTL-grounded learning strategies [45] as the operational implementation of these mechanisms within the REL-MIATTs framework (see Sections 9 and 10).

REL-MIATTs is applied to a bicycle lane segmentation task [46]. In this application, we treat ourselves as non-experts in identifying bicycle lanes in street images and assume that the true target of bicycle lanes does not exist in the real world for us. The results demonstrate the feasibility of Democratic Supervision and REL-MIATTs in supporting recognition, evaluation, and learning under the absence of an objectively accessible TT. Furthermore, the application suggests the potential of the framework for supporting individual education and professional development, consistent with prior discussions of Democratic Supervision in these domains [37–41]. The application further illustrates the broader implications of REL-MIATTs for continuous human–AI co-cognitive development and potential human–AI co-evolution. (see Section 11).

The contributions of this article are twofold: 1) it philosophically examines how shifts in assumptions regarding the existence and non-existence of the TT give rise to new perspectives and insights for ML-based predictive modelling; and 2) it correspondingly derives a knowledge system for recognition, evaluation and learning under Democratic Supervision for ML. The remainder of this article is structured as follows:

- Part 1 (Section 1) introduces the fundamental concepts of ML;
- Part 2 (Sections 2–3) philosophically examines the shifts in assumptions regarding the existence and non-existence of the TT, where Section 2 systematically analyses the existence assumption of the TT in mainstream ML paradigms, and Section 3 explicitly posits the non-existence assumption of the TT from the negative ontology perspective and under this assumption defines Democratic Supervision for ML;
- Part 3 (Sections 4–5) proposes a knowledge system under Democratic Supervision. Section 4 presents MIATTs as an instance-level implementation of Democratic Supervision for ML. Section 5 derives principles for recognition, evaluation, and learning with MIATTs; based on these principles, it establishes the REL-MIATTs framework and further characterizes REL-MIATTs from the perspective of Cognitive Machine Learning, discussing its potential implications for human–AI co-cognitive development and human–AI co-evolution;
- Part 4 (Sections 6–7) explores the theoretical implications and emerging applications of REL-MIATTs in rethinking machine learning foundations and advancing explanation discovery, causal exploration, scientific hypothesis generation, and human–AI collaborative knowledge evolution.
- Part 5 (Section 8) validates Human–AI Co-Evolution and Adaptive Knowledge Discovery in REL-MIATTs using a synthetic controlled environment.

- Part 6 (Sections 9–10) develops the algorithmic and operational foundations of REL-MIATTs, including generation and assessment of MIATTs, task-difficulty estimation, LAF-based evaluation, and UTTL-based learning strategies.
- Part 7 (Section 11) applies REL-MIATTs to a real-world bicycle lane segmentation task and demonstrates its feasibility as an implementation of Democratic Supervision, as well as its potential for supporting individual education and professional development;
- Part 7 (Sections 12-13) discusses and concludes the article.

# 1. Machine Learning

The objective of machine learning (ML) is to construct a predictive model with data collected for a specific prediction task based on efficient computing resources [47, 48]. This section introduces fundamental terminologies in ML, clarifies the interrelations among them, and discuss corresponding implications in shaping higher-level methodological dimensions.

Instance, label, and target constitute the fundamental components of the data associated with a particular ML task. To achieve the goal of ML in practical applications, evaluation and learning are two essential concepts, representing the core computational procedures required to develop an appropriate predictive model for the given task. Particularly, the evaluation procedure can be viewed as implicitly defining or inducing an optimization objective for learning procedure. Deployment in ML is the stage where a developed and assessed model is applied to unseen data in real-world settings. The deployed model can generate predictions and support decision-making, thereby operationalizing its learned capabilities beyond the evaluation and learning environment.

Collectively, these fundamental terminologies are the starting point for understanding how ML systems are structured. The interrelationships among them cover data collection, transformation, and solution implementation. Specifically, collected data consist of instances and labels, where labels are transformed into computational targets to enable model evaluation and learning. Based on the processed data, evaluation and learning are conducted in a coordinated manner, with evaluation providing essential support to the learning process. Through this interplay, a predictive model is constructed that can be deployed to map unseen instances to predicted targets, which can be further transformed back into interpretable labels. Section 1.4 illustrates the interrelationships among these core elements.

The six fundamental terminologies and the interrelations among them not only define how data is obtained and how predictive models are built, but also implicitly shape two higher-level methodological dimensions: evaluation paradigms and learning paradigms.

On one hand, evaluation paradigms focus on how the model performance is assessed, which is equally influenced by the nature of the available supervision. As will be discussed, the distinction between the provided accurate and inaccurate ground-truth labels (AGTLs and IAGTLs) gives rise to evaluation with AGTLs [49–53] and evaluation with IAGTLs [33–36], respectively. Thus, evaluation paradigms extend the same foundational considerations into the assessment stage, ensuring that model performance is interpreted appropriately under different data conditions.

On the other hand, learning paradigms are primarily concerned with how models are trained under different forms of supervision information derived from the data collection process. For example, the availability and quality of labels directly determine whether a task falls into unsupervised [54–56], supervised [1, 57, 58], weakly supervised [59–62], or reinforcement [63–66] learning. In this sense, learning paradigms operationalize the “solution implementation” aspect of the foundational terms by specifying how knowledge is extracted from data.

The evaluation and learning paradigms are intrinsically coupled: the choice of an evaluation paradigm often constrains or informs the appropriate learning paradigm. Together, they can be seen as systematic extensions of the fundamental terms. Evaluation paradigms govern how model performances are validated, and learning paradigms govern how models are built from data. They concretize the conceptual foundation for the full lifecycle of ML. Consequently, we formalize this coupling between evaluation and learning in the following proposition.

**Proposition 1.1 (Coupling of Evaluation and Learning in ML).** *Let an evaluation procedure be defined by a performance metric. Then, under standard machine learning formulations, the evaluation procedure can be viewed as implicitly defining or inducing an optimization objective for learning procedure, either directly or via surrogate losses.*

A systematic review providing further details on the topics covered in this section is available at https://doi.org/10.5281/zenodo.20173175.

# 2. Existence Assumptions of True Target in Current Mainstream Machine Learning Paradigms

Prior works [9, 13] have systematically examined the existence assumptions about TT underlying current major evaluation and learning paradigms. The evaluation paradigms considered include those based on accurate true targets (ATTs) [49–53] and those based on inaccurate true targets (IATTs) [13, 33–36], while the learning paradigms cover unsupervised learning [54–56], supervised learning (SL) [1, 57, 58], weakly supervised learning (WSL) [59–62], and reinforcement learning (RL) [63–66]. The key findings are summarized as Table 1.

While the prevailing existence assumption of TT across major evaluation and learning paradigms is that the TT objectively exists in the real world, a sceptical perspective questions this assumption by considering the possibility that such objective TT may not exist.

A systematic review providing further details on the topics covered in this section is available at https://doi.org/10.5281/zenodo.20373270.

# 3. Explicitly Posited Non-Existence Assumption of True Target and Defined Democratic Supervision for Machine Learning

We explicitly posit, in Assumption 3.1, the non-existence of the true target for ML from the perspective of Negative Ontology [9, 13]. The operationalized principles and corresponding applications across disciplines of Negative Ontology are presented in Supplementary, which together provide the scientific and philosophical foundations for this research.

**Assumption 3.1 (Negative Oncology of True Target for ML):** *The true target does not objectively exist in the real world as a universally accessible object.*

Under this non-existence assumption of TT, we define Democratic Supervision for ML by Definition 3.1. While the term "Democratic Supervision" has been discussed in contexts such as education [37–40] and professional development [41], here we introduce it into ML and provide a formal definition grounded in the non-existence assumption of TT. Democratic Supervision for ML is a supervision paradigm that, under the non-existence of a single true target, constructs supervisory signals through the aggregation, negotiation, or coexistence of diverse contributors' perspectives, rather than relying on a unique authoritative ground truth.

**Definition 3.1 (Democratic Supervision for ML).** *Under the non-existence assumption of true target, Democratic Supervision refers to a supervision paradigm in which the supervisory signals used for data preparation, evaluation, and learning are collaboratively generated by a diverse set of contributors, including domain experts, machine learning experts, and non-experts, without assuming the existence of an authoritative objectively correct ground truth. In this paradigm, supervision is constructed through the aggregation, negotiation, or coexistence of multiple perspectives, where each contribution reflects partial, context-dependent, or complementary aspects of the underlying reality. Consequently, Democratic Supervision replaces hierarchical, authority-driven labelling processes with a participatory and pluralistic framework, enabling models to be evaluated against and learn from supervision that embodies a distribution of viewpoints rather than a well-defined true target.*

Table 1. Summarization for existence assumptions of TT in learning paradigms [9]

| **Evaluation and Learning Paradigms** | | **Existence assumptions of TT** |
|---|---|---|
| **ATT-based evaluation** | Evaluation with massive ATTs | Both ATT-based and IATT-based evaluations fundamentally assume that well-defined TT objectively exists, enabling either direct use or indirect estimation despite inaccurate annotations [33–36, 49–53] |
| | Evaluation with limited ATTs | |
| **IATT-based evaluation** | Probabilistic selection methods | |
| | Error-bounded methods | |
| **Unsupervised learning** | | None: the concept of TT generally does not apply [54–56] |
| **Supervised learning** | Precisely supervised learning | The TT objectively exists in the real world [1, 57, 58] |
| | Moderately supervised learning | |
| | Hybrid forms that combine both | |
| **Weakly supervised learning** | Incomplete supervision | A prevailing mainstream TT assumption is observed in WSL-related surveys: the TT objectively exists in the real world, even though it may be partially missing, coarsely represented, or inaccurately observed in the available annotations [2, 4–8, 59–62, 67–74] |
| | Inexact supervision | |
| | Inaccurate supervision | |
| | Cross-scenario | |
| **Two prevalent paradigms in inaccurate supervision** | Learning from noisy labels (LNL) | Emerging trend of TT assumption is observed in more recent works in LNL [7, 8] and LMA [2–6]: researchers are beginning to question the clarity, definability, and uniqueness of TT in both noisy-label and multiple-annotator settings; and the shift in LMA appears to be more radical [3, 7, 8, 10–32] |
| | Learning from multiple annotators (LMA) | |
| **Reinforcement learning** | | Not specifically discussed, can be referred to the above assumptions |

Based on Proposition 1.1 (Coupling of Evaluation and Learning) and Definition 3.1 (Democratic Supervision for ML), we introduce the following proposition for the coupling of evaluation and learning under Democratic Supervision.

**Proposition 3.1 (Coupling of Evaluation and Learning under Democratic Supervision).** *Under the non-existence assumption of a true target, let the evaluation procedure be defined over a set or distribution of supervisory signals collaboratively generated by multiple contributors, as specified in Definition 3.1. Then, under standard machine learning formulations, the evaluation procedure induces an optimization objective for the learning procedure that reflects the aggregation, negotiation, or coexistence of these multiple perspectives, either directly or via surrogate losses. Consequently, the learning process is inherently coupled with a pluralistic evaluation procedure, such that the learned model optimizes with respect to a distribution of viewpoints rather than a unique true target objective.*

The definition of Democratic Supervision in ML suggests that new data preparation strategies and new evaluation and learning paradigms should be both explored. Together, these two explorations expand the scope toward evaluation and learning under Democratic Supervision. Under the conventional existence assumption, ML relies on domain expert-centric supervision, where paradigm design is largely constrained by the quality of domain expert-annotated data. In contrast, the non-existence assumption reconceptualizes supervision as a democratic process involving domain experts, ML practitioners, and potentially non-experts, enabling more flexible and inclusive data construction. This perspective generalizes the conventional setting and opens a broader space for developing evaluation and learning paradigms under democratic/pluralistic supervision.

The fundamental shift in the assumption of TT, moving from the conventional existence assumption (or the sceptical "may-not-exist" view) to the explicitly posited non-existence assumption, leads to substantially different perspectives and insights on supervision and ML paradigms, as summarized in Table 2. These new perspectives and insights lead to evaluation and learning under Democratic Supervision that transcend the boundaries between ATT-based and IATT-based evaluation, as well as between SL-based and WSL-based learning, unifying them within a generalized paradigm.

From a philosophical standpoint, the non-existence assumption departs from traditional views by rejecting the necessity of an authority-defined TT, thereby enabling new forms of supervision grounded in democratic participation. In contrast to conventional ML paradigms, the non-existence assumption supports supervision structures that are more inclusive and participatory. This transition not only reshapes how supervision is conceptualized but also introduces significant transformations and advantages at both the humanistic level, by promoting the democracy of expertise, and the technical level, by motivating more natural approaches to data preparation, evaluation, and learning.

Further details on the topics discussed in this section are available at https://doi.org/10.5281/zenodo.20422058.

## 4. Presented Multiple Inaccurate True Targets as an Instance-Level Realization of Democratic Supervision

Grounded in the non-existence assumption of TT, Democratic Supervision enables a more inclusive research landscape, thereby extending ML research toward evaluation and learning under such a paradigm. This section presents a component for operationalizing Democratic Supervision at the instance level through Multiple Inaccurate True Targets (MIATTs).

Table 2. Shifts in assumptions on TT existence and their implied perspectives and insights for supervision democracy and ML paradigms [9]

| Abbreviation of TT assumption | Interpretation regarding existence of TT | Implied perspectives and insights | Supervision democracy | Evaluation and learning paradigms |
|---|---|---|---|---|
| Existence assumption of TT (Or may-not-exist assumption of TT) | TT objectively exists in the real world (Or TT may not objectively exist in the real world) | Domain experts lead the supervision process, while ML experts play a supporting role and non-experts are excluded | Domain Expert-Centric Supervision | ATTs/IATTs-based evaluation and SL/WSL-based learning |
| Non-existence assumption of TT | TT does not objectively exist in the real world as a universally accessible object | Even if a TT physically exists (e.g., in conventional SL settings), one can still deliberately adopt this assumption; Experts (domain and ML experts) and non-experts can equally and collaboratively lead the supervision process; | Democratic Supervision | Evaluation and learning under Democratic Supervision |

In this section, we present Definition 4.1 and corresponding Remark 4.1, which formalize MIATTs as a finite set of inaccurate true targets [43].

**Definition 4.1 (Multiple Inaccurate True Targets, MIATTs).** *Let $t^*$ serve as a latent conceptual reference for the underlying (undefinable) target of a given machine learning, and let $SF(t^*)$ denote the complete set of semantic facts that, as a latent formal reference, would characterize $t^*$. Consistent with Assumption 3.1, $t^*$ and $SF(t^*)$ are not assumed to exist as universally accessible supervisory objects and are not directly available. A MIATTs set associated with $t^*$ is a finite collection*

$$MIATTs = \{t_n{}^* | n \in \{1, \cdots, N\}\}, \;\; N \geq 2$$

*where each $t_n{}^*$ is an inaccurate true target satisfying:*

***1) Partial representation:***

$$SF(t_n{}^*) \subset SF(t^*),$$

*i.e, each $t_n{}^*$ encodes only a subset of the underlying true target's semantic facts.*

***2) Collective coverage:***

$$\cup_{n=1}^{N} SF(t_n{}^*) \subseteq SF(t^*),$$

*with the possibility that $\cup_{n=1}^{N} SF(t_n{}^*) = SF(t^*)$.*

*In other words, no single $t_n{}^*$ fully specifies $t^*$, but together the MIATTs set captures one or more of its essential aspects.*

**Remark 4.1 (Latent Reference).** The use of $t^*$ and $SF(t^*)$ *serve only as latent conceptual and formal references but not as ontological commitments for defining the semantic relationship between MIATTs and the underlying (undefinable) target. They are not provided to, queried by, or estimated by the proposed algorithms. Consequently, the use of $t^*$ and*

*$SF(t^*)$ in formal definitions and theoretical analysis does not imply access to a true-target oracle. All operational procedures rely exclusively on the available MIATTs and task observations.*

Building on this definition, MIATTs is an instance-level realization of Democratic Supervision, grounding the abstract paradigm in a concrete supervisory structure. MIATTs enables the decomposition of evaluation and learning under Democratic Supervision into two executable components: evaluation over a set of inaccurate targets and learning under the uncertainty induced by this set. Within the MIATTs framework, evaluation and learning remain inherently coupled, as the way multiple targets are aggregated directly determines the learning objective. Finally, evaluation and learning with MIATTs together constitute an instance-level realization of evaluation and learning under Democratic Supervision. These are formally expressed as Propositions 4.1-4.4.

**Proposition 4.1 (MIATTs as an Instance-Level Realization of Democratic Supervision).** *Under the non-existence assumption of the true target, the multiple inaccurate true targets (MIATTs) framework constitutes an instance-level realization of Democratic Supervision.*

**Proposition 4.2 (From MIATTs to Executable Evaluation and Learning Procedures under Democratic Supervision).** *Under the non-existence assumption of the true target, the MIATTs representation enables the decomposition of evaluation and learning procedures under Democratic Supervision into two executable components: (i) evaluation with MIATTs (a set of inaccurate but collectively informative targets), and (ii) learning with MIATTs (target uncertainty induced by the set structure).*

**Proposition 4.3 (Coupling of Evaluation and Learning with MIATTs).** *Under the non-existence assumption of the true target, let supervision for each instance be represented by a MIATTs set $MIATTs = \{{t_n}^* | n \in \{1, \cdots, N\}\}$ as defined in Definition 4.1. Then, any evaluation procedure defined over MIATTs (i.e., over a set of partial and collectively informative targets) induces a corresponding learning objective that depends on how these targets are aggregated or jointly considered. Consequently, evaluation and learning are inherently coupled through the MIATTs structure, such that the learned model optimizes with respect to multiple inaccurate targets rather than a unique true target.*

**Proposition 4.4 (Evaluation and Learning with MIATTs as an Instance-Level Realization of Evaluation and Learning under Democratic Supervision).** *Under the non-existence assumption of the true target, evaluation and learning procedures defined with MIATTs constitute an instance-level realization of evaluation and learning under Democratic Supervision. Specifically, when supervision is represented by MIATTs at the instance level, both evaluation and learning are operationalized through set-based targets and their induced uncertainty, while preserving the pluralistic and coupled nature of Democratic Supervision.*

Overall, MIATTs is a unifying bridge between the conceptual paradigm of Democratic Supervision and its concrete, executable realization in machine learning systems. Further details on the topics discussed in this section are available at https://doi.org/10.5281/zenodo.20683453.

# 5. Proposed REL-MIATTs: Recognition, Evaluation and Learning

## with Multiple Inaccurate True Targets

Building upon MIATTs, in this section, we propose the REL-MIATTs framework for recognition, evaluation and learning with MIATTs [43] under Democratic Supervision to enable potential continuous human-AI co-evolution. The framework is grounded in logic-driven generation and assessment (LDGA) for MIATTs construction (recognition with MIATTs) [42], logical assessment formula (LAF) for evaluation with MIATTs [13], undefinable true target learning (UTTL) for learning with MIATTs [9], and empirical validation of principles of LDGA, LAF and UTTL through real-world applications. We further characterize REL-MIATTs from the perspective of Cognitive Machine Learning [44] and discusses its potential for supporting human–AI co-cognitive development toward potential human-AI co-evolution.

## 5.1 LDGA for MIATTs construction (recognition with MIATTs)

Definition 4.1 defines MIATTs as a finite set of partial semantic targets that jointly approximate an underlying true target for Democratic Supervision. In this context, their generation and assessment are crucial as they directly determine the quality and structural validity of the supervisory signal. As implied by Proposition 4.1 and Proposition 4.2, this step is essential for transforming Democratic Supervision from an abstract concept into an operational multi-target framework. Accordingly, based on Definition 4.1 and foundational concepts in logic [75, 76], particularly abductive reasoning [77] and Boolean algebra [78], this subsection presents the formation of LDGA and its practical principle for MIATTs construction. Intrinsically, LDGA-based MIATTs construction is formulated as a process for recognition with MIATTs under the absence of an objectively accessible TT.

### 5.1.1 Formation of LDGA

Based on Definition 4.1 and foundational concepts in logic, particularly abductive reasoning and Boolean algebra, referring to prior works [10–12, 79], the formation of LDGA is formally denoted as

$$LDGA\begin{cases} input: d \\ PC\begin{cases} G_{MIATTs} = ARGenerate(d; p^{ARG}) \\ Q_{MIATTs} = BAAssess(G_{MIATTs}; p^{BAA}) \end{cases} \\ output: (G_{MIATTs} = \{t_1{}^*, \cdots, t_N{}^*\}, Q_{MIATTs} = score \in [0,1]) \end{cases}. \quad (1)$$

The input of LDGA is an instance ($d$) captured from a finite world.

LDGA constitutes of two processing components ($PC$), including abductive reasoning-based generation and Boolean algebra-based assessment:

1) Generating MIATTs ($G_{MIATTs}$) from the input instance $d$, the abductive reasoning-based generation component is formally expressed as

$$G_{MIATTs} = ARGenerate(d; p^{ARG}) = \{t_1{}^*, \cdots, t_N{}^*\}. \quad (2)$$

2) Assessing the quality of the generated MIATTs $G_{MIATTs}$, the Boolean algebra-based assessment component if formally expressed as

$$Q_{MIATTs} = BAAssess(G_{MIATTs}; p^{BAA}) = score \in [0,1]. \quad (3)$$

Each $p$ of expressions (2)-(3) denotes the hyper-parameters corresponding to the implementation of respective expression. $G_{MIATTs}$ is a set of multiple inaccurate true targets associated with $d$ and $Q_{MIATTs}$ indicates the quality score of $Q_{MIATTs}$.

Finally, the output of LAF is the pair result $(G_{MIATTs} = \{t_1{}^*, \cdots, t_N{}^*\}, Q_{MIATTs} = score \in [0,1])$.

### 5.1.2 Principle of LDGA

Based on the formulation of LDGA, the practical principle of LDGA for MIATTs construction (recognition with MIATTs) can be summarized as Principle 5.1.

**Principle 5.1 (Practicability of LDGA for MIATTs construction (Recognition with MIATTs))**.

*Grounded in Logic, LDGA enables systematic MIATTs construction.*

*Abductive reasoning enables distributed and multi-perspective MIATTs generation by allowing both experts and non-experts to contribute valid explanatory hypotheses, thereby realizing Democratic Supervision.*

*Boolean algebra enables formal validity assessment and quantitative quality evaluation of MIATTs through logical consistency and collective coverage analysis.*

*LDGA characterizes prediction difficulty from the perspective of supervision generation, where the generation of MIATTs captures the complexity of generating semantic explanations and the assessment of MIATTs captures their reliability as approximations of the inaccessible true target.*

*These mechanisms of LDGA establish a theoretically grounded and computationally executable framework for constructing MIATTs under Democratic Supervision.*

## 5.2 LAF for evaluation with MIATTs

Proposition 4.2 shows that MIATTs enable executable evaluation procedures under Democratic Supervision. Evaluation with MIATTs is important because it replaces objective evaluation with a structured multi-perspective (democratic) framework. Instead of measuring error against an inaccessible ground truth, models are assessed against a set of partial yet complementary targets. Accordingly, grounded in MIATTs, together with a series of logical reasoning processes under uncertainty [80–84], this subsection derives the formation and usage of Logical Assessment Formula (LAF), LAF-based model performance evaluation and the practical principle of LAF for evaluation with MIATTs [13].

### 5.2.1 Formation of LAF

Grounded in MIATTs and reasoning under uncertainty, according to prior work [13], the formation of LAF is formally denoted as

$$LAF\begin{cases} inputs: \begin{cases} t \\ MIATTs = \{t_1{}^*, \cdots, t_N{}^*\} \end{cases} \\ PC \begin{cases} LF = LogicalFactNarrate(MIATTs; p^{LFN}) \\ LC = LogicalConsistencyEstimate(t, LF; p^{LCE}) \\ LAM = LogicalAssessmentMetricBuild(LC; p^{LAM}) \end{cases} \\ output: LAM = \{LAM_1, \cdots, LAM_w\} \end{cases} . \quad (4)$$

The inputs of LAF include the predicted true target ($t$) and the MIATTs ($MIATTs = \{t_1{}^*, \cdots, t_N{}^*\}$). The inputs $t$ and $MIATTs$ are corresponding to the same underlying true target associated with an instance.

LAF constitutes of three processing components ($PC$), including logical fact narration, logical consistency estimation and logical assessment metric build:

1) Narrating the logical facts ($LF$) from the $MIATTs$, the logical fact narration component is formally expressed as

$$LF = LogicalFactNarrate(MIATTs; p^{LFN}) = \{LF_1, \cdots, LF_f\}. \quad (5)$$

2) Estimating the logical consistencies ($LC$) between the predicted $t$ and the narrated $LF$, the logical consistency estimation component is formally expressed as

$$LC = LogicalConsistencyEstimate(t, LF; p^{LCE}) = \{LC_1, \cdots, LC_u\}. \quad (6)$$

3) Producing the logical assessment metrics ($LAM$) based on the estimated $LC$, the logical assessment metric build component is formally expressed as

$$LAM = LogicalAssessmentMetricBuild(LC; p^{LAM}) = \{LAM_1, \cdots, LAM_w\}. \quad (7)$$

Each $p$ of expressions (6)-(8) denotes the hyper-parameters corresponding to the implementation of respective expression. The narrated $LF$ is a list of qualitative descriptions that logically represent the facts contained in the $MIATTs$, the estimated $LC$ is also a list of qualitative descriptions that logically represent the consistencies between the predicted $t$ and the narrated $LF$, and the built $LAM$ is a series of abstractly formalized metrics that are derived from the qualitative descriptions of the estimated $LC$ to represent the discrepancies between the predicted $t$ compared with the underlying true target.

Finally, the output of LAF is a series of logical assessment metrics ( $LAM = \{LAM_1, \cdots, LAM_w\}$) that can reflect certain discrepancies of the predicted $t$ compared with the underlying true target.

The LAF-based model performance evaluation (LAF-MPE) procedure can be established to assess how a given model performs on a specific prediction task. The input of LAF-MPE is a set of logical assessment metrics ($LAM$) generated by the model-specific LAF, and the output is a quantitative logical model performance (LMP) normalized to the range [0,1], reflecting the logical performances of the model on the given task. The LAF-MPE strategy can be formally expressed as

$$LMP = LAF_{MPE}(LAM; p^{LAF_{MPE}}) = quant \in [0,1]. \quad (8)$$

Here, $p^{LAF_{MPE}}$ denotes the hyperparameters governing the implementation of Formula

### 5.2.2 Principle of LAF for evaluation with MIATTs

Based on the MIATTs-based formation and usage of LAF, the practical principle of LAF for evaluation with MIATTs can be summarized as Principle 5.2.

**Principle 5.2 (Practicability of LAF for Evaluation with MIATTs)**. *Grounded in MIATTs, LAF enables a logical evaluation framework that approximates conventional evaluation based on accurate true targets without requiring direct access to the underlying true target.*

*LAF can approximate true-target-based evaluation reasonably well in difficult tasks, while potentially exhibiting greater deviations in easier ones.*

*These mechanisms of LAF establish an executable evaluation procedure under Democratic Supervision, whose reliability is dependent on task difficulty.*

## 5.3 UTTL for learning with MIATTs

Proposition 4.2 shows that MIATTs enable executable learning procedures under Democratic Supervision, while Proposition 4.3 emphasizes the intrinsic coupling between evaluation and learning through the MIATTs structure. In this context, learning with MIATTs is important because it enables model optimization under multiple, potentially incomplete and diverse supervision signals rather than an objectively predefined target. Moreover, the coupling between evaluation and learning implies that LAF provides the evaluation signals required for subsequent learning with MIATTs, ensuring that model optimization remains consistently aligned with multi-perspective (democratic) performance feedback. Accordingly, grounded in MIATTs, LAF-based evaluation with MIATTs, and the objective optimization principle of machine learning [85, 86], this subsection derives the formulation and usage of Undefinable True Target Learning (UTTL) and establishes its practical principle for learning with MIATTs [9].

### 5.3.1 Formation of UTTL

Grounded in MIATTs, the objective optimization principle of machine learning and LAF-based evaluation with MIATTs, according to prior work [9], the formation of UTLL can be

formally denoted as

$$UTTL\begin{cases} inputs: \begin{cases} d \\ MIATTs = \{t_1{}^*, \cdots, t_N{}^*\} \\ f_\theta \\ LMP = LAF_{MPE}\left(LAF: PC\left(f_\theta(d), MIATTs; \overline{p^{LAF}}\right); \overline{p^{LAF_{MPE}}}\right) \end{cases} \\ PC = \begin{cases} l = loss(f_\theta(d), MIATTs; p^{loss}) \\ S_\theta = arg \underset{\theta \in \Theta}{\mathrm{opt}}(l; p^{opt}) \\ \theta^* = arg \underset{\theta \in S_\theta}{\max} LMP \end{cases} \\ output: f_{\theta^*} \end{cases}. \quad (9)$$

The inputs of UTTL include an input instance ($d$), its corresponding MIATTs ($MIATTs = \{t_1{}^*, \cdots, t_N{}^*\}$), a machine learning predictive model $f_\theta$ parameterized by $\theta$, and a LAF-based method performance evaluation (LAF-MPE) procedure $LMP = LAF_{MPE}\left(LAF: PC\left(f_\theta(d), MIATTs; \overline{p^{LAF}}\right); \overline{p^{LAF_{MPE}}}\right)$ parameterized by $\overline{p^{LAF}} = \left\{\overline{p^{LFN}}, \overline{p^{LCE}}, \overline{p^{LAM}}\right\}$ and $\overline{p^{LAF_{MPE}}}$.

Particularly, the predictive model $f_\theta$ can map the input instance $d$ into a predicted true target by $f_\theta(d)$; The $LAF_{MPE}$ procedure can transform the predicted true target $f_\theta(d)$ and corresponding $MIATTs$ into a quantitative logical method performance ($LMP = quant \in [0,1]$); The parameters of $LAF_{MPE}$ (i.e., $\overline{p^{LAF}} = \left\{\overline{p^{LFN}}, \overline{p^{LCE}}, \overline{p^{LAM}}\right\}$ and $\overline{p^{LAF_{MPE}}}$) are explicitly defined for LAF-related operations.

UTTL constitutes of three processing components ($PC$), including loss construction, loss-driven model optimization and LMP-driven model selection:

1) Based on the predicted true target $f_\theta(d)$ and corresponding $MIATTs$, a loss function needs to be constructed. This component is formally expressed as
$$l = loss(f_\theta(d), MIATTs; p^{loss}). \quad (10)$$
Here, $p^{loss}$ denotes the hyperparameters for implementing the loss function.
2) Optimizing the parameter $\theta$ of the model $f_\theta$ with respect to the loss function $l$, the model optimization component is formally expressed as
$$S_\theta = arg \underset{\theta \in \Theta}{\mathrm{opt}}(l; p^{opt}) = \{\theta_1, \cdots, \theta_S\}. \quad (11)$$
Here, $\Theta$ is the complete parameter space of $\theta$ and $p^{opt}$ denotes the hyperparameters for implementing the model optimization component, and $S_\theta$ is a set of $\theta$ parameters associated with the predictive model $f_\theta$.
3) Selecting the optimal parameter $\theta^*$ from $S_\theta$ that enables the predictive model $f_\theta$ obtain the max $LMP$ value, the model selection component is formally expressed as
$$\theta^* = arg \max_{\theta \in S_\theta} LMP. \quad (12)$$

Note that, by Formulas (11) and (12), the direct optimization of $f_\theta$ with respect to LMP is decomposed into loss-driven model optimization and LMP-driven model selection, enabling practical learning under MIATTs-based supervision when LAF-based evaluation signals are non-differentiable.

Finally, the output of UTTL is the optimized model $f_{\theta^*}$ which aims to uncover the underlying true target (i.e., the underlying structural reality) associated with the input instance $d$.

### 5.3.2 Principle of UTTL

Based on the LAF-based formation of UTTL, the practical principle of UTTL for learning with MIATTs can be summarized as Principle 5.3.

**Principle 5.3 (Practicability of UTTL for Learning with MIATTs)**. *Grounded in MIATTs and LAF-based evaluation, UTTL enables model optimization without direct access to the underlying true target by coupling learning objectives with logically derived evaluation signals.*

*UTTL can achieve performance reasonably close to conventional true-target-based learning for difficult tasks, where LAF provides reliable approximations of true-target evaluation. In easier tasks, where LAF may exhibit larger deviations from true-target-based evaluation, UTTL may correspondingly experience larger deviations in the optimized models.*

*UTTL enables practical optimization of MIATTs-based logical performance by decomposing direct LAF-based optimization into loss-driven model optimization and LAF-based model selection without changing the optimal solution.*

*These mechanisms of UTTL establish an executable learning procedure under Democratic Supervision, whose reliability is dependent on task difficulty.*

## 5.4 Empirical validation of Principles 5.1–5.3 through real-world applications

Principle 5.1 provides a theoretical foundation for MIATTs construction (recognition with MIATTs). Some prior works in the field of medical image analysis have explored LDGA-based approaches, particularly those based on abductive reasoning, for generating multiple semantic targets or hypotheses, providing empirical evidence supporting the practicability of Principle 5.1 [10–12].

Principle 5.2 provides a theoretical foundation for evaluation with MIATTs. Prior works in medical image analysis have developed a series of LAF-based evaluation algorithms, demonstrating the feasibility of logical assessment under MIATTs-based supervision and providing empirical evidence that the reliability of LAF-based evaluation depends on task complexity [13, 14].

Principle 5.3 provides a theoretical foundation for learning with MIATTs through UTTL. Prior works in medical and urban image analysis have developed UTTL-based learning strategies, providing empirical evidence supporting the practicability of optimizing predictive models under MIATTs-based supervision [9–12, 46].

Based on the above theoretical principles and their corresponding real-world applications, we draw the following Conclusion 5.1 regarding the empirical support for Principles 5.1–5.3.

**Conclusion 5.1 (Real-World Empirical Support for Principles 5.1–5.3)** *Existing real-world applications provide empirical support for Principles 5.1–5.3. Specifically, prior studies in medical and urban image analysis have demonstrated the feasibility of LDGA-based MIATTs construction, LAF-based evaluation with MIATTs, and UTTL-based learning with MIATTs. These empirical results suggest that Principles 5.1–5.3 are not merely theoretical derivations but practically applicable principles for developing algorithms in real-world prediction tasks.*

## 5.5 REL-MIATTs

Together, Assumption 3.1, Definitions 3.1–4.1, Principles 5.1–5.3 and Conclusion 5.1 provide the scientific foundation, as summarized in Table 3, for establishing the Recognition, Evaluation and Learning with MIATTs (REL-MIATTs) framework. Particularly, Proposition 4.4 indicates that REL-MIATTs is an Instance-Level realization of evaluation and learning framework under Democratic Supervision.

### 5.5.1 Formation of REL-MIATTs

Based on the formations and principles of LDGA, LAF and UTTL, according to prior work [43] the formation of REL-MIATTs can be formally denoted as

$$REL_{MIATTs}\begin{cases} inputs:\begin{cases} d \\ f_\theta \end{cases} \\ PC:\begin{cases} (G_{MIATTs}, Q_{MIATTs}) = LDGA:PC\left(d; \overline{p^{LDGA}}\right) \\ LMP = LAF_{MPE}\left(LAF:PC\left(f_\theta(d), G_{MIATTs}; \overline{p^{LAF}}\right); \overline{p^{LAF_{MPE}}}\right) \\ f_{\theta^*} = UTTL:PC\left(d, G_{MIATTs}, f_\theta, LMP; \overline{p^{UTTL}}\right) \\ \tilde{t}^* = f_{\theta^*}(d) \end{cases} \\ outputs: \{(G_{MIATTs}, Q_{MIATTs}), LMP, f_{\theta^*}, \tilde{t}^*\} \end{cases}. \quad (13)$$

The inputs of REL-MIATTs include an input instance ($d$) and a machine learning predictive model $f_\theta$ parameterized by $\theta$.

REL-MIATTs constitutes of four processing components ($PC$), including LDGA-based MIATTs construction, LAF-based method performance evaluation (LAF-MPE), UTTL-based learning, and optimized model deployment:

1) Grounded in the usage of LDGA, the LDGA-based MIATTs construction component ($LDGA:PC$) is parameterized by $\overline{p^{LDGA}} = \left\{\overline{p^{ARG}}, \overline{p^{BAA}}\right\}$, where these parameters are explicitly defined for LDGA-related operations. Given the input instance $d$ captured from a finite world, $LDGA:PC$ generates a set of multiple inaccurate true targets ($G_{MIATTs} = \{t_1{}^*, \cdots, t_N{}^*\}$) and its quality ($Q_{MIATTs} = score \in [0,1]$) to causally obtain various partial properties of the underlying structural reality under $d$.
2) Grounded in the usage of LAF, the LAF-MPE component ($LAF_{MPE}$) is parameterized by $\overline{p^{LAF}} = \left\{\overline{p^{LFN}}, \overline{p^{LCE}}, \overline{p^{LAM}}\right\}$ and $\overline{p^{LAF_{MPE}}}$, where these parameters are explicitly defined for LAF-related operations. Given the predicted true target $f_\theta(d)$ and the $G_{MIATTs}$ generated by $LDGA:PC$, $LAF_{MPE}$ generates a quantitative logical method performances ($LMP = quant \in [0,1]$) which can used for logical evaluations of the predictive model $f_\theta$.
3) Grounded in the usage of UTTL, the UTTL-based learning component ($UTTL:PC$) is parameterized by $\overline{p^{UTTL}} = \left\{\overline{p^{loss}}, \overline{p^{opt}}\right\}$, where these parameters are explicitly defined for UTTL-related operations. Given the input instance $d$, the $G_{MIATTs}$ generated by $LDGA:PC$, the predictive model $f_\theta$ and the logical method performance $LMP$ generated by $LAF_{MPE}$, $UTTL:PC$ produces a final optimized predictive model $f_{\theta^*}$.
4) The optimized model deployment component uncovers the structural reality $\tilde{t}^*$ underlying the input instance $d$ through the final optimized predictive model $f_{\theta^*}$, which is expressed as $\tilde{t}^* = f_{\theta^*}(d)$.

The outputs of REL-MIATTs include, but are not limited to, $\{(G_{MIATTs}, Q_{MIATTs}), LMP, f_{\theta^*}, \tilde{t}^*\}$, which are available for necessary user applications.

Table 3. Scientific foundation for REL-MIATTs

| Scientific foundation | | | | | | |
|---|---|---|---|---|---|---|
| Assumption 3.1 | Definitions | | Principles | | | **Proposal** |
| | Definition 3.1 | Definition 4.1 | Principle 5.1 | Principle 5.2 | Principle 5.3 | |
| Negative oncology of TT for ML | Democratic Supervision for ML | Multiple Inaccurate True Targets, | Practicability of LDGA (recognition with MIATTs) | Practicability of LAF (evaluation with MIATTs) | Practicability of UTTL (learning with MIATTs) | REL-MIATTs under Democratic |

| | | |
|---|---|---|
| MIATTs | | Supervision |
| | Conclusion 5.1 | |

### 5.5.2 Iteration of REL-MIATTs

Particularly, the framework parameters $\{\overline{p^{LDGA}}, \{\overline{p^{LAF}}, \overline{p^{LAF_{MPE}}}\}, \overline{p^{UTTL}}\}$ are not fixed after initialization. Instead, they can be iteratively analysed and modified by users based on domain knowledge, evaluation feedback and accumulated experience, which are gained through continuous interactions with the REL-MIATTs outputs $\{(G_{MIATTs}, Q_{MIATTs}), LMP, f_{\theta^*}, \tilde{t}^*\}$. This forms the iteration of REL-MIATTs: the system evolves iteratively by $REL_{MIATTs_1} \rightarrow REL_{MIATTs_2} \rightarrow \ldots$, which leads to $f_{\theta_1^*}/\tilde{t}_1^* \rightarrow f_{\theta_2^*}/\tilde{t}_2^* \rightarrow \ldots$ and simultaneously improvs human understanding of the underlying structural reality.

### 5.5.3 Human-centric principle of REL-MIATTs

Based on the formation of REL-MIATTs, the human-centric principle of REL-MIATTs can be summarized as Principle 5.4.

**Principle 5.4 (Human-Centric Practicability of REL-MIATTs)** *Grounded in LDGA-based MIATTs construction, LAF-based evaluation and UTTL-based learning, REL-MIATTs enables an open system, in which users can actively participate in the construction, evaluation, learning, and deployment processes of machine learning systems rather than only passively receiving model predictions.*

*The applicability of REL-MIATTs depends on prediction difficulty characterized by MIATTs-based difficulty representation, making REL-MIATTs particularly suitable for difficult tasks where accurate true targets are inaccessible or undefinable.*

*By requiring only basic task-relevant knowledge, REL-MIATTs enables users to interact with MIATTs-based outputs, expand their understanding of underlying structural reality, and iteratively improve both predictive models and the EL system itself. This supports self-education and long-term self-professional development.*

*These mechanisms of REL-MIATTs establish a human-centric AI framework under Democratic Supervision, in which users are transformed from passive consumers of predictions into active participants in knowledge acquisition, system refinement, and continuous human–AI co-evolution.*

## 5.6 Instantiation of REL-MIATTs

The instantiation of REL-MIATTs refers to the process of transforming the theoretical REL-MIATTs framework into an executable machine learning system for a specific real-world task. This process follows the four major components of REL-MIATTs and iteration: MIATTs construction, evaluation, learning, optimized model deployment, and human interaction and iterative refinement, which can be visually depicted as Fig. 1.

The complete instantiation process can be summarized as: $d, f_\theta \rightarrow LDGA \rightarrow (G_{MIATTs}, Q_{MIATTs}) \rightarrow LAF \rightarrow LMP \rightarrow UTTL \rightarrow f_{\theta^*} \rightarrow \tilde{t}^*$. Here, LDGA instantiates supervision construction through democratic MIATTs; LAF instantiates evaluation without direct access to true targets; UTTL instantiates learning under MIATTs-based supervision; and Human interaction instantiates continuous refinement and knowledge evolution. The instantiation of REL-MIATTs converts the theoretical principles of Democratic Supervision into an executable, adaptive, and human-centric machine learning system.

## 5.7 REL-MIATTs and Cognitive Machine Learning: Toward human–AI co-

## evolution

Cognitive Machine Learning (CML) has been proposed as an approach that combines machine learning with cognitive mechanisms, with the aim of developing learning systems capable of more cognitively sophisticated forms of learning [44]. The relationship between REL-MIATTs and CML can be understood through a progression from their different theoretical starting points to their potential implications for human–AI co-evolution. First, they originate from different theoretical starting points: CML primarily approaches machine learning from the perspective of incorporating cognitive mechanisms, whereas REL-MIATTs starts from the ontology and supervision of the true target. Second, the CML perspective provides a basis for characterizing the cognitive properties of REL-MIATTs, particularly in terms of the emergence, complementarity, and evolution of learning. Third, when iterative REL-MIATTs is extended to human–AI interaction, it provides a basis for human–AI co-cognitive development, in which human and AI cognitive states can mutually influence subsequent learning. Fourth, accumulated co-cognitive development may provide a potential basis for human–AI co-evolution. Overall, this perspective positions REL-MIATTs as a different route toward CML, extending machine learning beyond predefined-target learning toward continuous cognitive development without an objectively accessible TT.

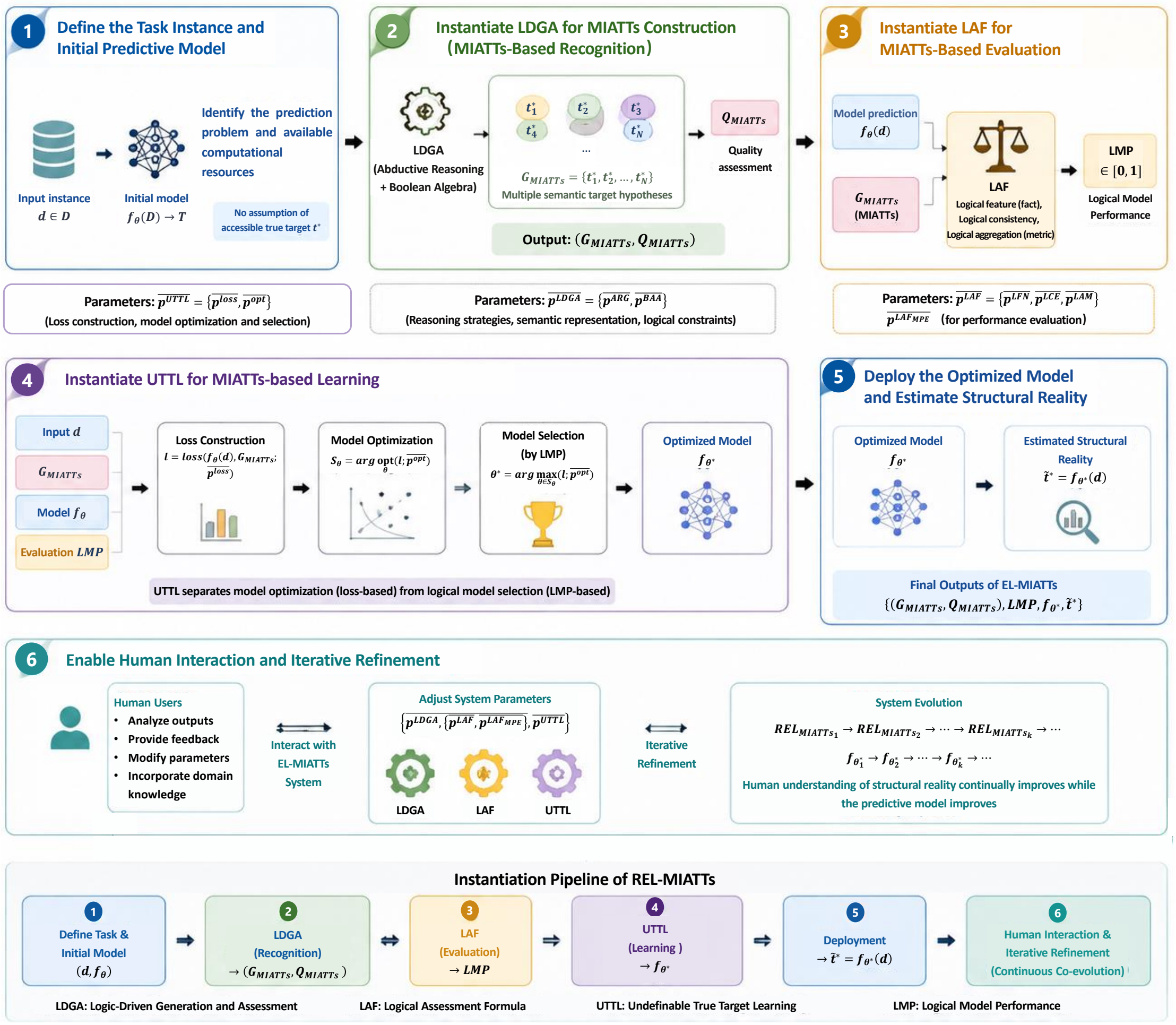

Figure. 1. Visualized instantiation pipeline of REL-MIATTs.

Further details on the topics discussed in this section are available at https://doi.org/10.5281/zenodo.21427125.

# 6 Theoretical Foundations and Implications of REL-MIATTs

The development of REL-MIATTs is motivated not only by the need to address machine learning problems where true targets are inaccessible, undefined, or inherently ambiguous, but also by a broader reconsideration of the epistemological foundations, human–AI relationships, and scientific implications of machine learning. By introducing MIATTs-based supervision, logic-driven evaluation, and uncertainty-aware learning, REL-MIATTs and its iteration extend machine learning beyond target-oriented prediction toward adaptive exploration and refinement of possible representations of underlying structural realities.

In this section, we further analyse REL-MIATTs from three complementary perspectives. The first perspective examines why REL-MIATTs is fundamentally different from traditional machine learning. Specifically, it investigates how REL-MIATTs differs from existing machine learning paradigms in terms of ontological assumptions about targets, supervision mechanisms, evaluation principles, learning objectives, and the role of human participants.

The second perspective examines what new capabilities REL-MIATTs introduces beyond traditional machine learning and who can benefit from these capabilities in the era of increasingly powerful AI. This perspective analyses how REL-MIATTs transforms AI from a prediction-oriented computational tool into a human-centred interactive framework that enables users, including non-experts, to participate in knowledge exploration through structured reasoning, uncertainty assessment, and logical evaluation. Such a transformation provides new opportunities for human knowledge acquisition, self-education, professional development, and continuous human–AI co-evolution, while potentially reducing knowledge inequality caused by increasingly complex AI systems.

The third perspective explores why REL-MIATTs matters beyond conventional machine learning. By examining its epistemological, scientific, and causal implications, this perspective investigates the potential of REL-MIATTs as a framework for adaptive truth exploration, AI-assisted scientific discovery, and human–AI collaborative exploration of underlying realities.

Together, these discussions position REL-MIATTs not merely as a machine learning framework for improving predictive performance, but as a potential emerging paradigm for human-centred adaptive knowledge discovery. In this paradigm, AI and humans jointly participate in the continuous construction, evaluation, and refinement of knowledge, enabling a new form of human–AI co-evolution under conditions of uncertainty and incomplete understanding. Further details on the topics discussed in this section are available at https://doi.org/10.5281/zenodo.21486635.

# 7 Applications and Future Potential of REL-MIATTs: Toward Human-Centred Adaptive Knowledge Discovery

The primary objective of machine learning has traditionally been to develop computational models capable of accurately mapping observed data to predefined targets. This paradigm has achieved remarkable success in numerous applications, particularly when reliable supervision is available and the target concept is sufficiently well-defined. However, many real-world problems do not satisfy these assumptions. In complex scientific, medical, engineering, and social domains, observations often represent only partial manifestations of deeper and potentially unknown structural realities. In such scenarios, the central challenge is not merely predicting an outcome, but discovering possible explanations, mechanisms, and causal structures underlying observed phenomena.

REL-MIATTs is proposed under this broader perspective. By replacing the assumption of directly accessible true targets with the concept of multiple inaccurate true targets (MIATTs), REL-MIATTs provides a framework for exploring inaccessible structural realities through logic-driven hypothesis construction, logical evaluation, and adaptive learning. Therefore, the potential applications of REL-MIATTs extend beyond conventional predictive modelling toward explanation discovery, causal exploration, scientific hypothesis generation, and human–AI collaborative knowledge evolution.

In this section, we discuss the potential applications of REL-MIATTs from several interconnected perspectives:

1. REL-MIATTs as an adaptive explanation discovery framework;
2. REL-MIATTs as an AI-assisted hypothesis exploration system;
3. REL-MIATTs as an abductive causal exploration framework;
4. REL-MIATTs as a foundation for AI-assisted scientific discovery;
5. REL-MIATTs as a human-centred education and professional development framework.

These perspectives reflect that REL-MIATTs has the potential to transform AI systems from prediction-oriented tools into adaptive partners for exploring uncertain or unknown realities. Further details on the topics discussed in this section are available at https://doi.org/10.5281/zenodo.21820040.

# 8 Validating Human–AI Co-Evolution and Adaptive Knowledge Discovery in REL-MIATTs

The comparison between REL-MIATTs and Cognitive Machine Learning in Section 5.7 suggests that REL-MIATTs provides a potential basis for Human–AI Co-Evolution. Furthermore, the theoretical implications and emerging applications discussed in Sections 6 and 7 position REL-MIATTs as a potential emerging paradigm for human-centred adaptive knowledge discovery, with potential applications in explanation discovery, causal exploration, scientific hypothesis generation, and collaborative Human–AI knowledge evolution. Accordingly, this section validates the potential for Human–AI Co-Evolution and Adaptive Knowledge Discovery within the REL-MIATTs framework using a synthetic controlled environment (SCE). We focus particularly on the implementation of REL-MIATTs in the SCE and analyse how its iterative recognition, evaluation, and learning processes contribute to Human–AI Co-Evolution and Adaptive Knowledge Discovery. With successive mutual Human-AI Co-Evolution, the REL-MIATTs system eventually ‘jumps’ to be close to the explanatory structure that does not objectively exist in the SCE as a universally accessible object. This controlled validation provides a reference for understanding and potentially applying REL-MIATTs in real-world scenarios.

## 8.1 Implementation of REL-MIATTs in SCE

In the SCE, REL-MIATTs is implemented as a four-stage iterative process progressing from the non-expert to the more advanced professional level. At each stage, a stage-specific subset of the observable phenomenon $D_s$ is used to construct multiple inaccurate and potentially complementary MIATTs through controlled stochastic sampling around the latent explanatory structure $T$, with progressively smaller deviations and fewer MIATTs as the knowledge state develops. The generated MIATTs are then used to evaluate and optimize the machine learning model through a simplified least-squares aggregation, yielding an updated AI learning state. The resulting AI behaviours provide potential patterns or explanatory structures that can support subsequent human recognition, while Human remains responsible for determining whether these candidates constitute new knowledge. The updated Human–AI knowledge state is then carried forward to the next stage, forming the iterative process $State_s \rightarrow Recognition_s(MIATTs\ Construction) \rightarrow Evaluation_s \rightarrow Learning_s \rightarrow State_{s+1} \rightarrow Recognition_{s+1}$. In this way, the synthetic implementation preserves the functional structure of REL-MIATTs while abstracting away the implementation-specific complexity of LDGA and LAF, thereby enabling controlled analysis of Human–AI Co-Evolution and Adaptive Knowledge Discovery.

## 8.2 Experimental results and analysis

The experimental results of REL-MIATTs across the four progressive stages, the non-expert (NE/s1), semi-professional (Spro/s2), professional (Pro/s3), and more advanced professional (MAPro/s4) levels, are presented in Fig.2, Fig. 3 and Table 4. Specifically, Fig. 2 primarily presents the results of Human-to-AI enhancement; Fig. 3 presents the potential AI-to-Human enhancement and the corresponding REL processes; and Table 4 summarizes the Human–AI Co-Evolution, including the knowledge states, process measurements, and stage-wise changes in the integrated knowledge state throughout the four-stage REL-MIATTs process.

First, the results demonstrate Human–AI Co-Evolution through measurable Human-to-AI enhancement and qualitatively analysed potential AI-to-Human enhancement. Second, they show that MIATTs disagreement can function as an informative and generative mechanism by preserving alternative explanatory perspectives, expanding the space available for human recognition, and supporting progressive model development. Third, the stage-wise evolution of the knowledge state demonstrates adaptive Recognition, Evaluation, and Learning, forming a closed-loop process in which knowledge acquired at one stage influences subsequent stages. Finally, the evolving knowledge states and their correspondence with the inaccessible latent structure provide evidence for Adaptive Knowledge Discovery. Together, the analyses establish the progression: from Human–AI Co-Evolution to Informative MIATTs Disagreement to Adaptive REL to Adaptive Knowledge Discovery, while also identifying the qualitative cognitive jump through which Human-recognized MIATTs and AI-learned patterns may provide Human with new opportunities to recognize previously unconsidered explanatory structures.

## 8.3 Discussion

The synthetic results support REL-MIATTs as a Human–AI co-evolutionary framework rather than merely an iterative machine learning approach. Human-recognized MIATTs enhance AI learning, while improved AI behaviours can provide potential patterns that enhance subsequent Human recognition; MIATTs disagreement further serves as an informational and generative mechanism that preserves alternative explanatory possibilities. Through the resulting adaptive Recognition–Evaluation–Learning cycle, knowledge states progressively evolve and support Adaptive Knowledge Discovery. Although the synthetic implementation abstracts away the full complexity of LDGA and LAF, it provides evidence for the behavioural and evolutionary principles of REL-MIATTs. More broadly, Human–AI Co-Evolution suggests a shift from data-centric learning toward models shaped by evolving human knowledge, with potential applications in personal Human–AI co-evolutionary knowledge systems for continuous learning and knowledge development.

## 8.4 Conclusion

This validation used a synthetic controlled environment to examine the core behavioral and evolutionary properties of REL-MIATTs. The synthetic experiment provides evidence not only for the observed performance changes within REL-MIATTs, but also for the broader theoretical implications of its evolving Recognition–Evaluation–Learning process. In particular, the results suggest that REL-MIATTs should be understood as an evolving Human–AI knowledge process rather than as a conventional iterative machine learning procedure. The results support a progressive mechanism in which multiple inaccurate perspectives preserve informative disagreement, enabling mutual Human–AI adaptation and continuous knowledge

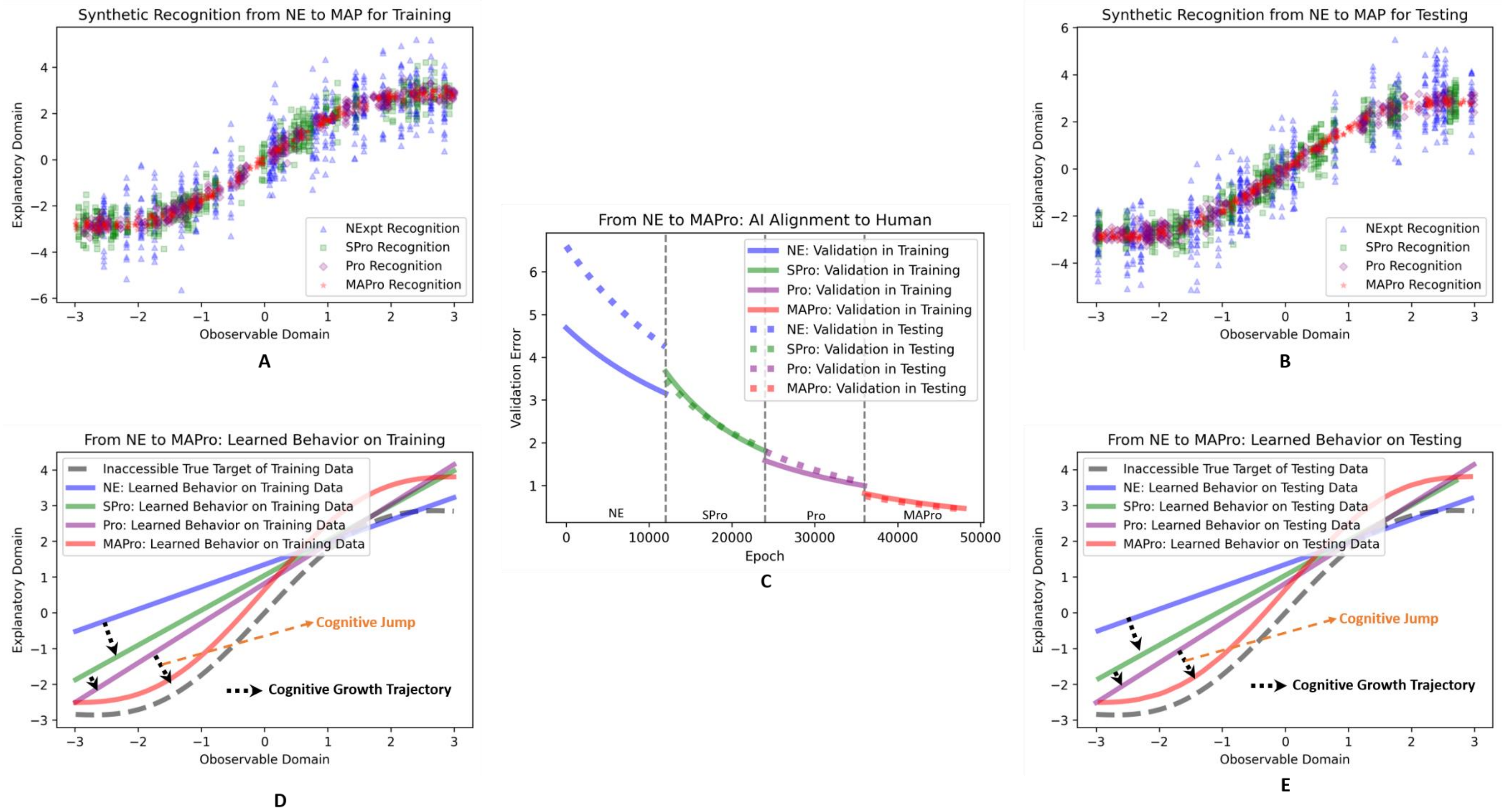

Figure 2. Human-to-AI enhancement across the four progressive stages of REL-MIATTs. The figure shows the synthetic human-recognized MIATTs (A and B) together with their enabled corresponding changes in AI predictive capability from the NE (s1) to the MAPro (s4) stage (C, D and E).

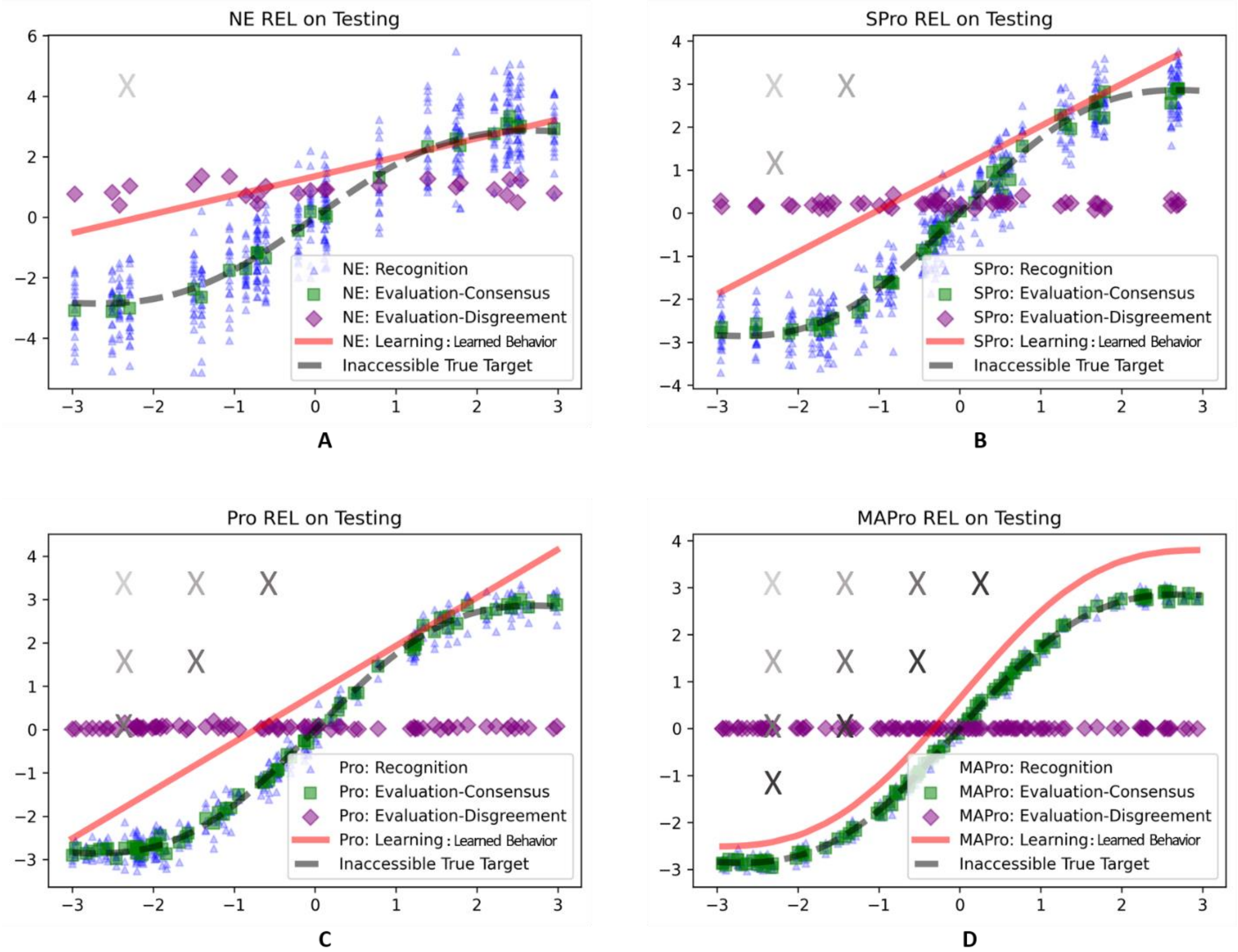

Figure 3. Potential AI-to-Human enhancement and the iterative REL processes across the four progressive stages of REL-MIATTs. The figure illustrates how the improved predictive capability of the optimized AI model may reveal previously unrecognized patterns or explanatory structures (X or jumping from linear (C) to non-linear (D)), which can provide opportunities for subsequent human recognition and knowledge refinement through the iterative recognition–evaluation–learning process.

Table 4. **Human–AI Co-Evolution and stage-wise knowledge-state evolution in REL-MIATTs.** The table summarizes the Human and AI knowledge states, process measurements, and stage-wise changes in the integrated Human–AI knowledge state across the four progressive stages, highlighting the iterative mutual enhancement underlying the REL-MIATTs process.

| **REL-MIATTs Processes** | **NE (s1)** | | **SPro (s2)** | | **Pro (s3)** | | **MAPro (s4)** | |
|---|---|---|---|---|---|---|---|---|
| **State** | $State_{s1}$: | $H_{s1}! = \emptyset$<br>$AI_{s1}=\emptyset$ | $State_{s2}$: | $H_{s2} = H_{s1} \cup C_{s1}^{H}$<br>$AI_{s2} = AI_{s1} \cup C_{s1}^{AI}$ | $State_{s3}$: | $H_{s3} = H_{s2} \cup C_{s2}^{H}$<br>$AI_{s3} = AI_{s2} \cup C_{s2}^{AI}$ | $State_{s4}$: | $H_{s4} = H_{s3} \cup C_{s3}^{H}$<br>$AI_{s4} = AI_{s3} \cup C_{s3}^{AI}$ |
| **Inputs** | $Inputs_{s1}$: | $D_{s1}$<br>$f_{\theta_{s1}}: f_{\theta_{s1}}(d) = w_{s1}d + b_{s1}$ | $Inputs_{s2}$: | $D_{s2}$<br>$f_{\theta_{s2}}: f_{\theta_{s1}^{*}}$ | $Inputs_{s3}$: | $D_{s3}$<br>$f_{\theta_{s3}}: f_{\theta_{s2}^{*}}$ | $Inputs_{s4}$: | $D_{s4}$<br>$f_{\theta_{s4}}: f_{\theta_{s3}^{*}}(d) + sin\ (d)$ |
| **REL** | $REL_{s1}$: | $R_{s1}$: $MIATTs_{s1}$<br>$E_{s1}$: $Con_{s1}^{MIATTs}$<br>$E_{s1}$: $Dis_{s1}^{MIATTs}$<br>$L_{s1}$: $f_{\theta_{s1}^{*}}: f_{\theta_{s1}^{*}}(d) = w_{s1}^{*}d + b_{s1}^{*}$ | $REL_{s2}$: | $R_{s2}$: $MIATTs_{s2}$<br>$E_{s2}$: $Con_{s2}^{MIATTs}$<br>$E_{s2}$: $Dis_{s2}^{MIATTs}$<br>$L_{s2}$: $f_{\theta_{s2}^{*}}$ | $REL_{s3}$: | $R_{s3}$: $MIATTs_{s3}$<br>$E_{s3}$: $Con_{s3}^{MIATTs}$<br>$E_{s3}$: $Dis_{s3}^{MIATTs}$<br>$L_{s3}$: $f_{\theta_{s3}^{*}}$ | $REL_{s4}$: | $R_{s4}$: $MIATTs_{s4}$<br>$E_{s4}$: $Con_{s4}^{MIATTs}$<br>$E_{s4}$: $Dis_{s4}^{MIATTs}$<br>$L_{s4}$: $f_{\theta_{s4}^{*}}$ |
| **H-to-AI Enhancement** | $C_{s1}^{AI}$: | $f_{\theta_{s1}^{*}} \backslash f_{\theta_{s1}}$ | $C_{s2}^{AI}$: | $f_{\theta_{s2}^{*}} \backslash f_{\theta_{s2}}$ | $C_{s3}^{AI}$: | $f_{\theta_{s3}^{*}} \backslash f_{\theta_{s3}}$ | $C_{s4}^{AI}$: | $f_{\theta_{s4}^{*}} \backslash f_{\theta_{s4}}$ |
| **AI-to-H Enhancement** | $C_{s1}^{H}$: | $K_{s1}$: Fig.3.A(X) | $C_{s2}^{H}$: | $K_{s2}$: Fig.3.B(X) | $C_{s3}^{H}$: | $K_{s3}$: Fig.3.C(X)<br>$K_{s3}^{+}$ : Explanatory structure is probably non-linear and sin-function-like (**Cognitive Jump**) | $C_{s4}^{H}$: | $K_{s4}$: Fig.3.D(X)<br>$K_{s4}^{+}$ : Explanatory structure is largely sin-function-like |
| **Outputs** | $Outputs_{s1}$: | $f_{\theta_{s1}^{*}}$ and etc.⋯ | $Outputs_{s2}$: | $f_{\theta_{s2}^{*}}$ and etc.⋯ | $Outputs_{s3}$: | $f_{\theta_{s3}^{*}}$ and etc.⋯ | $Outputs_{s4}$: | $f_{\theta_{s4}^{*}}$ and etc.⋯ |

-state evolution. The evolving knowledge state subsequently provides an adaptive basis for recognition, evaluation, and learning, ultimately supporting Adaptive Knowledge Discovery. These position REL-MIATTs as a potential application framework for personalized education, professional development, and lifelong learning.

The synthetic environment provides controlled evidence that REL-MIATTs can support Human–AI Co-Evolution and that Adaptive Knowledge Discovery can emerge from an evolving Recognition–Evaluation–Learning process rather than from fixed-target optimization alone. It also provides a conceptual reference for extending REL-MIATTs to real-world MIATTs construction, where domain-specific recognition, heterogeneous supervisory perspectives, and more complex evaluation mechanisms can replace the simplified synthetic implementations. However, the current validation does not directly reproduce the complexity of real-world MIATTs construction or LAF for evaluation with MIATTs, nor does it involve real human participants or real-world domain knowledge. In addition, the progressive development of human recognition is partly predefined by the synthetic experimental design, and therefore the AI-to-Human enhancement observed here is interpreted primarily as qualitative evidence of potential cognitive support rather than a direct causal measurement. These limitations motivate future validation with real human participants, domain-specific MIATTs construction, and the full logical evaluation mechanisms of REL-MIATTs.

Further details on the topics discussed in this section are available at https://doi.org/10.5281/zenodo.22254777.

# 9 LDGA-Based Algorithms for MIATTs construction and MIATTs-Based Algorithm for Task-Difficulty Estimation

The principles of LDGA establish the necessity of developing concrete algorithms for generating and assessing MIATTs and for estimating task difficulty based on MIATTs. Specifically, the MIATTs construction is a fundamental prerequisite for realizing REL-MIATTs, because MIATTs serve as the core supervisory structures that replace inaccessible true targets under the negative ontology assumption. Without an effective mechanism for generating and assessing MIATTs, democratic supervision and MIATTs-based evaluation and learning cannot be operationalized. Furthermore, the MIATTs-based task-difficulty estimation is essential for determining the applicability of REL-MIATTs across different prediction tasks. By quantitatively characterizing task difficulty from the perspective of supervision generation, this algorithm provides a criterion for identifying scenarios where REL-MIATTs can provide advantages over conventional true-target-based learning. Therefore, this section presents the LDGA-based algorithms for MIATTs generation and assessment [42], together with the MIATTs-based algorithm for task-difficulty estimation, to help transforming the processing components of REL-MIATTs into executable ones. Further details on the topics discussed in this section are available at https://doi.org/10.5281/zenodo.21839973.

# 10 LAF-Grounded Evaluation Strategies and UTTL-Grounded Learning Strategies under MIATTs-Based Supervision

While REL-MIATTs establishes a principled basis for addressing epistemic uncertainty, its practical implementation, which extends beyond the generation and assessment of MIATTs, critically depends on two complementary mechanisms that operationalize the framework in real-world ML tasks: LAF-based evaluation and UTTL-based learning with MIATTs. The former enables logically consistent assessment of models when multiple partially correct targets coexist, while the latter allows model training to proceed effectively even when the true target is fundamentally undefinable. Together, these mechanisms transform REL-MIATTs from a conceptual framework into a deployable methodology for complex, real-world machine learning tasks. Their integration is thus crucial for achieving both theoretical rigor and practical utility in uncertain supervision settings. Accordingly, this section proposes applicable LAF-grounded evaluation and UTTL-grounded learning strategies as the operational realization of these two mechanisms within the REL-MIATTs framework [45]. Further details on the topics discussed in this section are available at https://doi.org/10.5281/zenodo.21863540.

# 11 Conducted Application of REL-MIATTs for Supporting Education and Professional Development for Individuals

Based on prior works [9, 13, 42, 43, 45], REL-MIATTs has been applied in bicycle lane segmentation task [46]. ***In this application, we treated ourselves as the non-expert at identifying bicycle lane in street images (i.e. assuming the TT of bicycle lane does not objectively exist in the real world as a universally accessible object for us). This application is conducted to demonstrate the potential of REL-MIATTs in supporting education and professional development for individuals.***

## 10.1 Implementation of REL-MIATTs

The implementation of REL-MIATTs framework for the bicycle lane segmentation can be summarized as Fig. 4. More details for this implementation are provided in [46].

The MIATTs generation for a given street image is implemented in three steps: Step 1—conceptualization of bicycle lanes using SAM; Step 2—annotation of inaccurate ground-truth labels using PixLab; Step 3—generation and assessment of MIATTs using a custom-developed executive program. For model development, we adopt the U-Net [87] architecture as the model for predicting bicycle lane in street images. The valuation metrics are established based on the LAF and common image segmentation metrics to quantify discrepancies between the model's predicted true target and the MIATTs (see Table 5). Guided by the established LAF-based metrics and the multi-target learning procedure within UTTL, the model undergoes iterative evaluation and learning using the MIATTs, culminating in the selection of the final optimized U-Net model. The optimized U-Net model can be deployed to predict the distribution over multiple plausible properties of the underlying true target, from which the desired true target can be derived.

## 10.2 Quantitative validation: Diverse metrics across epochs

To identify the final optimized task-specific U-Net model, we examined the temporal evolution of several key performance indicators—namely logical intersection-over-union (LIoU), logical error counts (LErrors, computed as LFP + LFN), and loss—across the optimization epochs, as depicted in Fig. 5. This visualization provides an intuitive overview of the model's learning dynamics, highlighting both the rapid performance gains in the early and mid-stages of optimizing and the gradual stabilization of metrics as convergence is approached. The trends in Fig. 5 also serve to pinpoint the epoch at which the model meets the predefined stopping criteria (LIoU > 0.999 and LErrors < 100), ensuring the selected model achieves both high accuracy and low error rates while avoiding unnecessary overtraining.

The trends depicted in Fig. 5 show the LIoU score has a consistent upward trajectory, rising from approximately 0.25 at epoch 20 to surpassing 0.999 after epoch 500. In parallel, the loss decreases steadily from about 0.70 to roughly 0.20, indicating progressive improvements in model fit. Total logical error counts (LErrors = LFP + LFN) remain above 1,300 during the early training stages but decline sharply after ~400 epochs, reaching a minimum of 53 at epoch 620. This point precisely satisfies the predefined stopping criteria (LIoU > 0.999, LErrors < 100), marking it as the optimal checkpoint. Selection of this epoch is further supported by the observed convergence pattern: performance metrics plateau beyond epoch 620, and logical error counts exhibit slight upward fluctuations thereafter, suggesting diminishing returns from continued training and a risk of overfitting. Thus, epoch 620 represents not only a quantitative optimum but also a strategically stable choice for balancing performance and generalization.

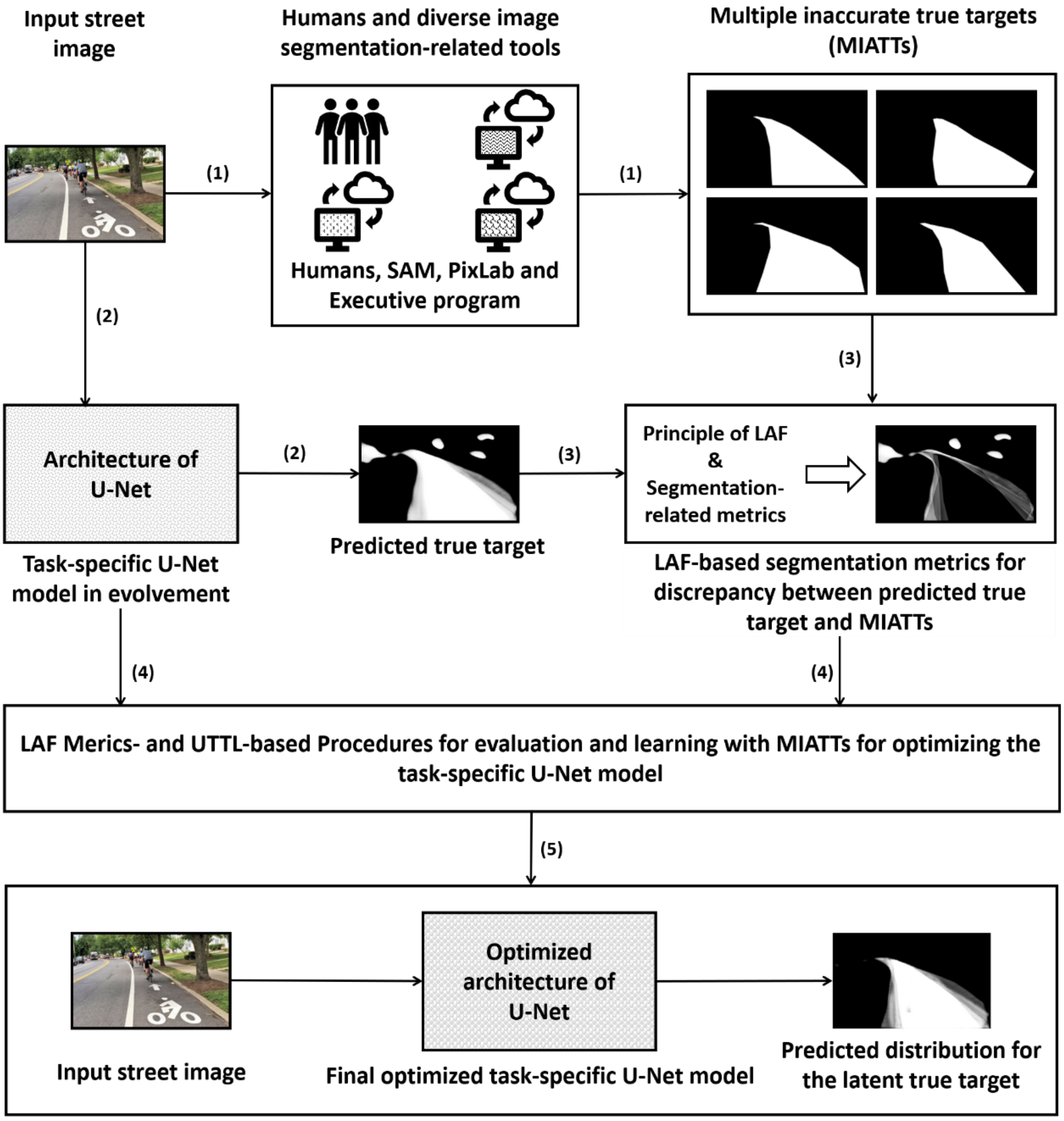

Figure 4. Visual summary of the REL-MIATTs framework implementation for the bicycle lane segmentation task on street images [46]. (1) MIATTs Generation; (2) Model Construction; (3) Metric Formulation; (4) Model Optimization; and (5) Deployment. The input street image was photographed by Chris Roell and is used here with acknowledgment.

Table 5. Summary of LAF-based metrics for image semantic segmentation [46]

| LAF-based metrics for image semantic segmentation | | | |
|---|---|---|---|
| **Logical TP (LTP)** | A pixel targeted as 'Logical object' is correctly predicted as 'Logical object' | **Logical Precision (LPrecision)** | LTP / (LTP+LFP) |
| **Logical FP (LFP)** | A pixel targeted as 'Logical non-object' is incorrectly predicted as 'Logical object' | **Logical Recall (LRecall)** | LTP / (LTP+LFN) |
| **Logical TN (LTN)** | A pixel targeted as 'Logical non-object' is correctly predicted as 'Logical non-object' | **Logical F1 (LF1)** | 2(LPrecision*LRecall) / (LPrecision+LRecall) |
| **Logical FN (LFN)** | A pixel targeted as 'Logical object' is incorrectly predicted as 'Logical non-object' | **Logical Accuracy (LAccuracy)** | (LTP+LTN) / (LTP+LFP+LTN+LFN) |
| **Logical Errors (LErrors)** | LFP + LFN | **Logical IoU (LIoU)** | LTP / (LTP+LFP+LFN) |

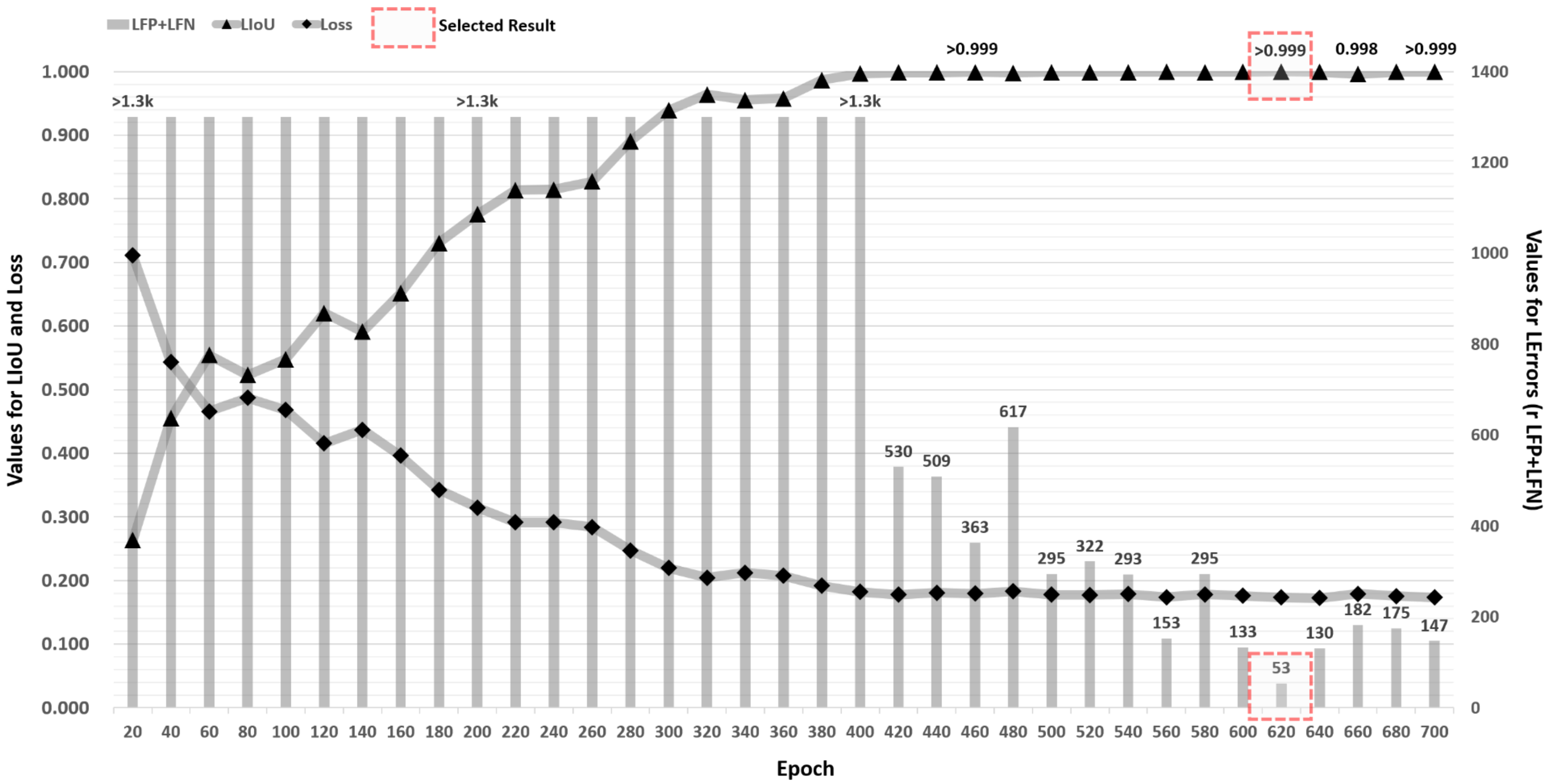

Figure 5. Evolution of LIoU, LErrors (LFP+LFN) and Loss over optimizing epochs [46]. Values included by red bounding boxes represent the best performance across all epochs. Evolution of key performance indicators across learning epochs for the task-specific U-Net model, including LIoU (black triangles), Loss (black circles), and total LErrors (LFP+LFN, grey bars). The stop criteria (LIoU > 0.999 and LErrors < 100) were first satisfied at epoch 620 (highlighted), yielding LIoU = 1.0 and only 53 errors, which was selected as the final optimized model. Early epochs show rapid gains in LIoU and steep reductions in Loss, while errors remain high until mid-learning. After around 500 epochs, LIoU plateaus near perfection, but errors continue to drop until the optimal point at epoch 620, after which slight fluctuations indicate no further improvement.

## 10.3 Qualitative comparison: Uncovered underlying true target vs. MIATTs

The model's predicted latent true target is compared with the MIATTs corresponding to the input street image to offer direct visual evidence of its ability to uncover a reasonable and coherent representation of the underlying target. Fig. 6 illustrates the alignment between the predicted true target binarized from the distribution prediction for the latent true target and the logical true target (LTT) derived from the MIATTs, while Fig. 7 presents side-by-side comparisons between the predicted true target (binarized distribution prediction) and each individual inaccurate true target in the MIATTs set. Together, these visualizations are presented to show that the final model produces spatially plausible segmentations aligned with the logical constraints defined by the LTT.

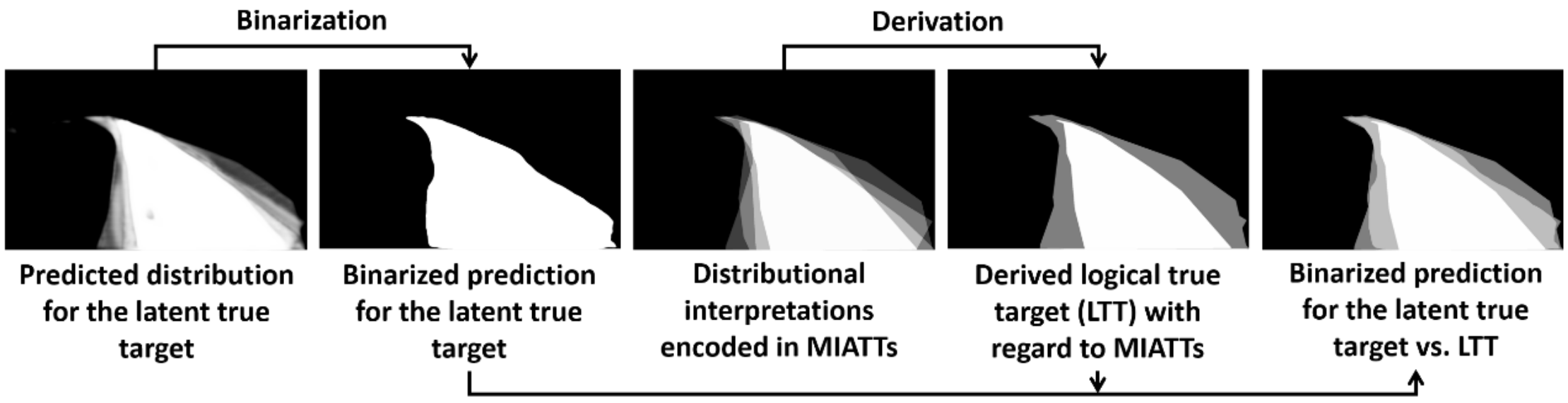

Figure 6. Visual comparison between the predicted true target binarized from the distribution prediction of the selected final optimized task-specific U-Net model for the latent true target and the logical true target (LTT) derived from the MIATTs [46]. The binarized prediction (threshold = 0.5) closely aligns with the LTT, showing that the model effectively constrains its segmentation within the logical boundaries inferred from multiple inaccurate true targets.

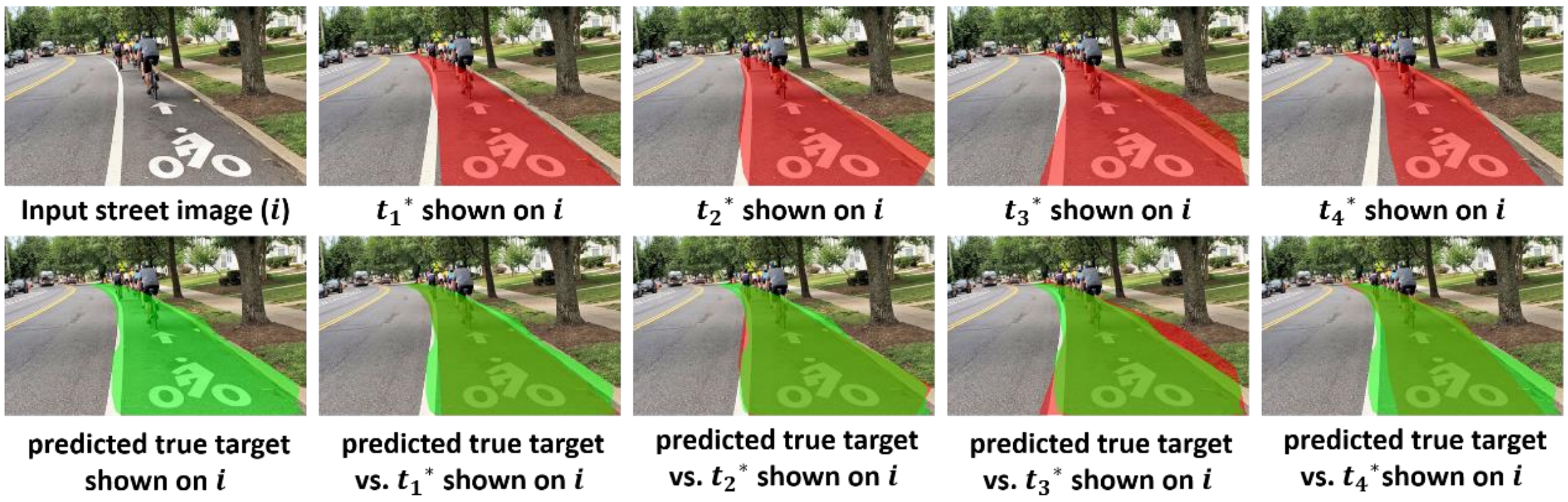

Figure 7. Visual comparison between the predicted latent true target of the selected final optimized task-specific U-Net model and each individual inaccurate true target within the MIATTs corresponding to the input street image [46]. The comparisons reveal that the model's output integrates complementary and consistent information from the MIATTs, while filtering out noise and biases present in individual inaccurate true targets. The input street image was photographed by Chris Roell and is used here with acknowledgment.

In Fig 6, the binarized prediction of the latent true target aligns closely with LTT derived from the MIATTs, indicating that the model has effectively learned to constrain its segmentation within the logically consistent boundaries inferred from multiple noisy and biased supervision sources. This demonstrates that the optimization process not only improved numerical performance but also yielded a prediction that is visually coherent with the consensus structure implied by the MIATTs. Fig. 7 further elucidates the interpretability of the model's output by comparing it against each individual inaccurate true target. The predicted segmentation captures the shared structural patterns present across the MIATTs while suppressing idiosyncratic errors, biases, or omissions found in individual targets. Together, these comparisons confirm that the model's output is not only statistically optimal according to the evaluation metrics but also visually and conceptually consistent with the underlying true target, thereby reinforcing confidence in its practical reasonability.

### 10.4 Potential of REL-MIATTs in supporting education and professional development for individuals

Within the REL-MIATTs framework, non-experts are not merely passive recipients of supervision but active participants in a learning process that unfolds alongside model development. As they construct predictive models, they are continuously exposed to comparisons between the model's predicted latent true target and the set of MIATTs associated with each input. This comparison provides direct and interpretable feedback: by observing how the model's predictions align with, reconcile, or differ from multiple inaccurate targets, non-experts are guided toward a more coherent understanding of the underlying (latent) true target. In this sense, the modelling process itself becomes an educational mechanism, gradually enhancing their cognition of the target concept.

Specifically, the model's predicted latent true target is compared with the MIATTs corresponding to each street image to offer visual evidence of how a reasonable and coherent representation of the underlying target can be inferred. As illustrated in Fig. 6, the binarized prediction derived from the distribution over the latent true target aligns closely with the logical true target constructed from the MIATTs. This alignment indicates that the model has successfully internalized the logical constraints implied by multiple, potentially inaccurate and biased supervision sources. Meanwhile, as provided in Fig. 7, the side-by-side comparisons between the predicted true target and each individual inaccurate true target, showing that the model captures shared structural patterns while suppressing individual inconsistencies, biases, or omissions. Through these visual and conceptual comparisons, non-experts can directly observe how disparate, inaccurate annotations collectively inform a more reliable representation.

Over time, as such educational interactions are repeated across many instances, non-experts progressively refine their understanding of the true target. What begins as instance-level insight accumulates into a more stable and transferable form of knowledge. When sustained over a longer period, this process effectively functions as professional training, enabling individuals to develop competencies related to true target cognition and task-specific judgment. In this way, REL-MIATTs supports not only immediate, self-guided education but also longer-term professional development.

This application thus demonstrates the potential of the REL-MIATTs framework to facilitate individuals' self-education and self-driven professional development. Such a role is consistent with prior discussions of Democratic Supervision in the fields of education and professional development [37–41], where it is understood as a collaborative and participatory approach in which authority is shared and stakeholders jointly engage in decision-making and reflective practice to support both collective and individual growth.

## 12 Discussion

In current mainstream machine learning paradigms, whether supervised learning, weakly supervised learning, noisy-label learning, or multi-annotator learning, there is an implicit shared assumption: the true target (TT) objectively exists in the real world, and our observations are merely imperfect. Even when dealing with uncertainty (e.g., noisy labels or multiple annotators), researchers often say that the TT "may not exist," but this does not truly deny the existence of the TT.

This article proposes a more radical philosophical stance: **the true target does not objectively exist in the real world as a universally accessible object** (Assumption 3.1). This is not an empirical conclusion but a deliberately chosen modelling assumption: even if the TT "seems to exist" in certain contexts, we can intentionally assume that it does not. The significance of this shift is profound: machine learning is no longer about "approximating the truth," but about "constructing actionable knowledge under conditions without a truth." In other words, the starting point of learning shifts from "existence" to "non-existence."

Under the non-existence assumption of TT, conventional learning with supervision immediately fails: there is no correct label, no authoritative annotator, and no absolute evaluation standard. To address this, this article proposes a new supervision paradigm: **Democratic Supervision** (Definition 3.1). Its core idea is that supervisory signals come from the aggregation, negotiation, and coexistence of perspectives from multiple stakeholders. Thus, domain experts are no longer "holders of the truth," non-experts can become valid information sources, and data itself becomes a "collection of perspectives." This implies a fundamental shift as shown in Table 6.

Table 6. Shift from Domain Expert-Centric Supervision to Democratic Supervision

| Domain Expert-Centric Supervision | Democratic Supervision |
|---|---|
| Domain Expert-driven | Multi-stakeholder participation |
| Authoritative ground truth | Participatory / Multiple perspectives |
| Standard answer | Semantic distribution |
| Static labels | Dynamic negotiation |

To translate this Negative Ontology of TT (Democratic Supervision) into computational structure for machine learning, a key bridge is introduced by Definition 3.2: **MIATTs (Multiple Inaccurate True Targets)**. Intuitively, a "true target" is no longer an authoritative point but a set composed of participatory / multiple perspectives. Each target is inaccurate, containing only a subset of semantic facts, but their union can reflect the hypotheses of the underlying structure. This brings an important shift: errors or inconsistencies are no longer a problem but a component of structure.

Based on MIATTs, a complete system is constructed: **REL-MIATTs (Recognition, Evaluation and Learning with MIATTs)**, which consists of three key layers:

1) **MIATTs construction (Recognition with MIATTs)**
   Using Logic-Driven Generation and Assessment (LDGA), MIATTs are constructed to have a recognition of the underlying true target.
2) **Evaluation (Evaluation with MIATTs)**
   Through the Logical Assessment Formula (LAF), evaluation no longer compares "prediction vs. ground truth," but "prediction vs. multiple inaccurate perspectives," transforming evaluation into a problem of logical consistency.
3) **Learning (Learning with MIATTs)**
   Through Undefinable True Target Learning (UTTL), the learning goal is no longer a definite label, but to minimize the "logical deviation" from multiple targets by LAF, turning learning into an optimization problem logically constrained by LAF-based metrics.

REL-MIATTs also provides a connection to Cognitive Machine Learning (CML) from a different theoretical starting point. While CML primarily focuses on incorporating cognitive mechanisms into machine learning, REL-MIATTs approaches cognitive learning from the Negative Ontology of true target and exhibits characteristics associated with the emergence, complementarity, and evolution of learning. A single REL-MIATTs cycle can be viewed as an elementary cognitive learning unit, while iterative REL-MIATTs enables continuous updating through feedback, experience accumulation, and state transformation. When extended to human–AI interaction, REL-MIATTs provides a basis for human–AI co-cognitive development, in which human and AI cognitive capabilities can mutually influence subsequent learning. This perspective suggests a possible transition from human-supervised learning toward human–AI co-evolution without requiring an objectively accessible TT.

Thus, in the REL-MIATTs framework, the model training process is simultaneously a process of human cognitive enhancement. For example, non-experts, by observing the relationship between model outputs and multiple annotations, gradually understand the "latent true structure" and ultimately form stable knowledge. This implies that **REL-MIATTs can function as a human–machine co-constructed cognitive system**.

The significance of this framework can be summarized at three levels:

1) **Ontological**: from "truly existing" → "structurally non-existent"
2) **Epistemological**: from "approximating truth" → "integrating perspectives"
3) **Methodological**: from "objective evaluation and learning" → "logical evaluation and learning"

The REL-MIATTs framework is particularly suitable for domains such as medical diagnosis (inherently ambiguous), subjective tasks (emotion, aesthetics), multi-annotator conflict scenarios, and problems without a standard answer. It provides a fundamentally new way of addressing the question: when "the answer does not exist," how can machine learning proceed?

This article **redefines the starting point of machine learning**. From: Ground Truth, Objective Target, Expert Authority → To: Non-existent Truth, Distributed Perspectives, Democratic Construction. Ultimately, it establishes a new knowledge system: machine learning starting from "existence" → machine learning starting from "non-existence."

## 13 Conclusion

In this article, philosophically examining the shifts in assumptions regarding the existence and non-existence of TT, we have shown that relaxing the existence assumption of TT to the non-existence assumption gives rise to a fundamentally different understanding of supervision. From the Negative Ontology of TT perspective, we explicitly posited that the TT does not objectively exist in the real world as a universally accessible object. Grounded in this non-existence assumption, we defined Democratic Supervision as a generalized supervision paradigm for ML that replaces reliance on an authoritative ground truth with the aggregation and coexistence of diverse, potentially inaccurate perspectives.

To operationalize Democratic Supervision, we presented Multiple Inaccurate True Targets (MIATTs) as its instance-level realization, and proposed a systematic knowledge system built upon MIATTs. Specifically, we established the logic-driven generation and assessment for MIATTs construction (recognition with MIATTs), developed a logical assessment formulation for evaluation with MIATTs, and introduced undefinable true target learning for learning with MIATTs. Together, these components establish the REL-MIATTs framework, which provides a coherent approach to predictive modelling under Democratic Supervision.

Through the conducted application to a bicycle lane segmentation task, we demonstrated that REL-MIATTs is not only capable of producing reasonable and interpretable predictions under multiple inaccurate supervision signals, but also serves as a mechanism for human-centred learning. In particular, the iterative comparison between model predictions and MIATTs enables non-experts to progressively refine their understanding of the undefinable true target, thereby supporting both self-directed education and longer-term professional development, which is aligning with prior discussions of democratic supervision in the fields of education and professional development. This highlights a broader implication of the framework: beyond improving modelling flexibility, it fosters a more inclusive and participatory interaction between humans and ML systems.

Overall, this article extends ML research in two key directions. First, it provides a philosophical reframing of supervision grounded in the non-existence assumption of the TT, offering new perspectives and insights into the nature of evaluation and learning. Second, it establishes a practical and theoretically grounded framework for conducting evaluation and learning under democratic/pluralistic supervision. Future work may further explore the scalability of MIATTs construction, the integration of richer forms of human input, and the application of REL-MIATTs to a wider range of domains where the true target is inherently ambiguous or contested.

# Supplementary: Negative Ontology as an Expansion Principle—Reframing Research and Practice via the Ontology of Non-Existence

Building on the ontology of non-existence, this supplementary proposes Negative Ontology as an Expansion Principle, a framework designed to extend the scope of research and practice by incorporating both fully describable dimensions ($D(X)$) and inherently excluded or undefined dimensions ($N(X)$). The principle is operationalized through three complementary mechanisms: Boundary Relaxation, Semantic Reversal, and Structural Reframing, which enable domain expansion, generative enhancement, and open-system superiority. Based on this, the Negative Ontology Reframing Methodology provides a systematic six-step approach for engaging with known and absent dimensions. Its application across Leadership, Information System, and Machine Learning demonstrates the framework's generalizability and capacity to extend theoretical, methodological, and practical boundaries. The framework challenges assumptions of computability, reductionism, and completeness, revealing the generative potential of absence, incompleteness, and uncertainty. This highlights that research and practice boundaries are defined not only by what is known but by the willingness to embrace the unknown, which is a perspective especially critical in the era of increasingly powerful AI.

## 1. Introduction

Modern scientific inquiry is largely grounded in the fundamental assumption that the world can be described, measured, and modelled. Across the natural sciences, social sciences, and engineering disciplines, research activities are typically organized around improving model accuracy, enhancing predictive capacity, and optimizing control mechanisms. This epistemic foundation is well supported in the philosophy of science literature, where positivist and empiricist traditions hold that legitimate scientific knowledge must be tied to observable, measurable phenomena and formal models [1, 2]. Thomas Kuhn's work also demonstrates that scientific practice is structured around paradigms, which are shared theoretical frameworks that presuppose the describability and structural regularity of phenomena, guiding how scientists conceptualize, measure, and interpret the world [3, 4]. More recently, philosophers of science have analysed the role of theory-laden observation and model-based reasoning in science, noting that the very structure of scientific practice privileges describable and predictable phenomena [5, 6]. Together, these works substantiate the claim that modern scientific practice is built upon the assumption of a describable and modellable world, linking philosophical justification with the practical orientation of contemporary research.

The emphasis on formal describability and predictability has driven unprecedented technological and empirical progress. However, critics have long pointed out that such emphasis also creates blind spots: it tends to underplay the role of context, lived experience, and emergent phenomena that resist neat codification. These blind spots are especially invisible to newcomers who are trained to work within established formal frameworks.

Philosophers and theorists have highlighted how scientific models often abstract away uncertainty and context to produce idealised representations, effectively excluding features that are complex, ambiguous, or hard to formalise [7, 8]. Phenomena such as tacit knowledge, situated practice, and affective experience challenge the assumption that everything relevant to science is describable in formal terms [9, 10]. The ontological commitment to definable, presentable, and codifiable entities underpins many mainstream research traditions [11, 12]. As critical realists point out, such commitments entail structural exclusions because they systematically ignore what cannot be fully expressed in formal models or measurement [13, 14]. These structural constraints shape what counts as valid data, legitimate explanation, and acceptable methodology in scientific practice.

In recent years, a trend has emerged across multiple fields, including Leadership, Information System, and Machine Learning: researchers have begun to focus on these "excluded dimensions." [15–17] From the perspective of Negative Ontology, these dimensions are not mere noise; instead, they are considered to have significant generative potential. This phenomenon suggests the need for a more foundational theoretical framework to explain why the inclusion of such negative dimensions expands the scope of research and practice. Accordingly, the central question of this article is: **Why does the introduction of the philosophy of Negative Ontology systematically expand the boundaries of research and practice across different domains?**

To address this question, this article starts from the ontology of non-existence and proposes a unified theoretical framework, **"Negative Ontology as an Expansion Principle"**, revealing its underlying mechanisms through formalization and cross-disciplinary analysis.

Specifically, we begin by formally defining Negative Ontology (Definition 1) and articulating its potential impact (Conclusion 1), after reviewing the concepts of Ontology and Negative Ontology in philosophy. Definition 1 shows that traditional (positive) Ontology can be understood as a special case of Negative Ontology, establishing the philosophical origin of this research. Conclusion 1 further reveals that the negative dimension is not merely an epistemic limitation but a source of generative potential, providing philosophical guidance for inquiry and practice. Together, they establish the ontology of non-existence for this study, emphasizing not merely what is known (existence) but, more importantly, what is absent or indescribable (non-existence).

Based on these foundations, we propose the Expansion Principle of Negative Ontology (Principle 1), which translates these philosophical insights into a formal and operationalizable framework. We establish the theoretical validity of Principle 1 and operationalize it through three complementary **Expansion Mechanisms**: **Boundary Relaxation**, **Semantic Reversal**, and **Structural Reframing**. Collectively, these mechanisms enable the systematic expansion of research and practice possibilities, extending both conceptual and practical horizons. Grounded in these mechanisms, we present the **Negative Ontology Reframing Methodology**, a six-step systematic approach for reframing the scope of inquiry. This methodology deliberately engages with both what is known and what is absent or indescribable, transforming traditionally excluded or marginal dimensions into productive sources of insight and action. The methodology's **application across Leadership, Information System, and Machine Learning** [15–17] demonstrates its generalizability and its capacity to extend inquiry beyond the limits of conventional, object-centered, or fully formalizable paradigms.

These contributions position Negative Ontology as an Expansion Principle as a framework with substantial theoretical significance, methodological innovation, and the potential to challenge entrenched assumptions of computability, strong reductionism, and completeness. This framework opens multiple avenues for future research, both in empirical hypothesis testing and in methodological and applied extensions. It fundamentally challenges the notion that the boundaries of research and practice are determined solely by what is already known. Instead, it establishes that **inquiry boundaries are shaped by our willingness to recognize, embrace, and work with the unknown, the uncertain, and the inexpressible, transforming absence and incompleteness into productive forces**. This insight is particularly critical in the era of increasingly powerful AI: **only by acknowledging that AI cannot encompass all the potential wisdom that future humans may generate, can we avoid a deterministic fate of replacement and continue to push the boundaries to create new forms of human intelligence that surpass AI**.

The remainder of this article is organized as follows:

- **Section 2:** Reviews the concepts of Ontology and Negative Ontology, and presents Definition 1 and Conclusion 1, which establish the ontology of non-existence for this study.
- **Section 3:** Proposes the Expansion Principle of Negative Ontology (Principle 1), discusses its theoretical validity, operationalizes it through three complementary Expansion Mechanisms: Boundary Relaxation, Semantic Reversal, and Structural Reframing, and discusses the possibilities of these mechanisms for systematic expansion of research and practice.
- **Section 4:** Presents the Negative Ontology Reframing Methodology, a six-step systematic approach that grounded in the three expansion mechanisms to engage both known and absent or indescribable dimensions.
- **Section 5:** Analyses the methodology's application across Leadership, Information System, and Machine Learning, demonstrating its generalizability and capacity to extend research and practice beyond conventional, object-centered paradigms.
- **Section 6:** Discusses theoretical and methodological contributions, highlighting the framework's potential to challenge assumptions of computability, reductionism, and completeness.
- **Section 7:** Outlines directions for future research, including empirical hypothesis testing, methodological extensions, and the development of Negative Ontology-based AI,
- **Section 8:** Concludes with reflections on the importance of embracing the unknown, particularly in the era of powerful AI.

## 2. Definition of Negative Ontology and Its Potential Impact

In this section, we begin by reviewing and summarizing the concepts of ontology and negative ontology in philosophy. We then present a formal definition of negative ontology that forms the philosophical origin premise and a formal conclusion regarding its potential impact on research and practice that forms the philosophical guidance. Together, the two establish the ontology of non-existence for this study.

## 2.1 Ontology

Ontology is the branch of philosophy concerned with the nature of being, existence, and the fundamental structure of reality, asking both what kinds of things exist and what it means for something to exist at all. In its classical formulation in Aristotle's Metaphysics [18], ontology is understood as the study of "being qua being," meaning that it investigates existence in the most general sense rather than focusing on any particular domain such as physics or biology.

In contemporary philosophy, ontology is typically treated as a core part of metaphysics and involves debates over what kinds of entities are fundamental, such as whether only physical objects exist (physicalism), whether abstract entities like numbers are real (Platonism), or whether many putative entities are merely linguistic constructs (nominalism). Quine [19] famously reformulated ontological commitment in terms of what a well-formed scientific theory must quantify over, suggesting that "to be is to be the value of a variable," thereby linking ontology closely to our best explanatory theories of the world.

More broadly, ontology also examines categories of being (objects, properties, events, relations), dependence structures among entities, and the criteria by which existence claims are justified [20, 21]. As such, ontology provides the conceptual foundation for much of philosophy and underlies how different disciplines implicitly define what kinds of entities their theories assume to exist.

In short, "Ontology" can be summarized as: The attempt to say what exists, and what it means for anything to exist at all."

## 2.2 Negative ontology

Negative ontology is not a standard school of thought rigorously established in metaphysics, but rather a philosophical approach that transcends multiple traditions. Its core lies not in defining "what exists" as in traditional ontology, but in approaching the meaning of "existence itself" through "negation," "absence," and "ineffability." In other words, it does not positively define existence, but reveals in what sense existence cannot be fully defined.

In traditional ontology, such as Aristotle's system [18], the task of philosophy is to classify and define existence: distinguishing between substance and attribute, seeking the essence of things, and describing it with clear concepts. However, negative ontology takes a reverse path: it points out that any definition inevitably excludes certain possibilities and is therefore always incomplete. Thus, existence itself is not fully presented in what we say, but rather reveals itself precisely in those parts that cannot be captured by language. As one scholar summarized: "Existence is not something that can be fully said, but rather appears at the limits of what can be said" (see Marion [22]).

This idea has deep roots in multiple philosophical traditions. First, in the "apophatic theology" of Christian thought, positivists like Pseudo-Dionysius [23] argue that God cannot be grasped through affirmative descriptions; rather, humanity can only approach divinity through continuous negation. For example, stating that "God is not finite," "not material," and "not within time". This approach actually reveals a deeper ontological view: the highest existence is precisely what cannot be conceptualized.

In 20th-century philosophy, this negative line of thinking was further developed. Martin Heidegger, in Being and Time, pointed out that the Western philosophical tradition has always focused on "beings," but forgotten "Being" itself [24]. He argued that Being is always both manifest and concealed; manifestation and concealment are inseparable. This "concealed structure" means that Being is never fully revealed, but always retains some unrevealable dimension [24].

Subsequently, Jacques Derrida further radicalized this line of thinking through the concept of "différance." In Of Grammatology, he argues that meaning does not originate from a stable presence, but rather from constant difference and delay. Therefore, any "existence" contains traces of absence; there is no final, fully present entity [25]. In this sense, existence itself is a process of constantly postponing completion.

Similar ideas can be found in Eastern philosophy. For example, Lao Tzu, in the Tao Te Ching, states, "The Tao that can be spoken of is not the eternal Tao," indicating that the true "Tao" cannot be fixed by language [26]. While the Buddhist concept of "emptiness" (śūnyatā) [27] emphasizes that all existence has no fixed, unchanging essence, but arises from relationships (Nāgārjuna, Mūlamadhyamakakārikā) [28]. These ideas all point in the same direction: existence is not a stable entity, but an open, incompletely definable structure.

Negative ontology can be understood through several key propositions: First, existence cannot be fully defined because any definition involves exclusion; second, "absence" is not simply non-existence, but rather an important part of the structure of existence; third, language inevitably fails in expressing existence because existence always transcends language; fourth, ontology is not a static entity, but a dynamic process, an unfolding of difference and tension.

An intuitive analogy is the "blank space" in painting. According to traditional ontology, we analyse the specific objects presented in the painting; however, from the perspective of negative ontology, what truly gives meaning to the painting is often the undepicted blank spaces. These "absent" elements are not insignificant, but rather constitute the conditions for the overall meaning (see François Jullien, The Great Image Has No Form [29]).

The importance of negative ontology lies in providing us with a way to understand experiences that cannot be directly expressed, such as the unspeakable in consciousness, the blank spaces in art, and existential anxiety and nothingness. It reminds us that reality is not a completely filled, fully explainable structure, but rather an open field composed of both "existence" and "absence."

In short, "Negative Ontology" can be summarized as: approaching "true existence" by revealing the "unspeakable parts." This approach does not deny existence itself, but rather rejects our overly certain understanding of existence, thus preserving a more open and profound dimension for philosophy.

### 2.3 Definition for negative ontology

Traditional ontology emphasizes the classification and definition of existence, aiming to grasp reality through a clear conceptual structure. In scientific methodology, this manifests as: clear definition of objects, measurability of variables, and model ability of systems. However, this approach implicitly includes an exclusionary mechanism: anything that doesn't meet these conditions is often considered noise or irrelevant.

In contrast, a key thread running through the history of philosophy emphasizes the role of "negativity." This idea holds that existence is not entirely presented in what can be expressed; the unspeakable is not blank but structural. In different traditions, this idea manifests as: existence is always accompanied by obscuration; meaning depends on difference and delay; formal systems exist as unprovable propositions. These ideas collectively point to a core viewpoint: any positive determination is inevitably incomplete.

Accordingly, we present the following Definition 1 regarding Negative Ontology:

**Definition 1 (Negative Ontology)**: *For any object $x$, its existence includes not only the describable part $D(x)$, but also the dimension $N(x)$ that cannot be fully expressed, and $N(x) \neq \emptyset$.*

**Definition 1 formally shows that traditional (positive) Ontology is a special case of Negative Ontology, which forms the philosophical origin for this study.**

## 2.4 Potential impact of negative ontology

If we consider "Negative Ontology" as a rigorous philosophical origin premise, its potential impact can be understood as the manifestation of the same structural proposition at different levels: any formalized, computational, or representational system cannot exhaustively encompass its objects, always leaving an irreducible "remainder." This "remainder" is precisely the negative dimension mentioned earlier, which is the part of existence that cannot be fully expressed or captured.

In the field of artificial intelligence, this idea directly challenges the fundamental assumption that "models can approximate reality." Since Alan Turing, computational theory has understood intelligence as a formalizable process, mapping the world to tractable representations through functions [30]. Modern machine learning has further developed into high-dimensional function approximation systems, using data to train models to capture the structure of the world. However, from the perspective of negative ontology, this mapping is incomplete in principle: the model learns only the "expressible part," failing to encompass the inexpressible dimensions of existence. Therefore, even under ideal conditions of infinite data and infinitely complex models, AI cannot achieve a complete grasp of the world.

This leads to several important consequences. First, AI's generalization ability has ontological limits, not just a data or algorithmic problem. Second, any representation system inevitably compresses differences, resulting in "semantic collapse," echoing the structure revealed by Jacques Derrida in his theory of "différance": meaning depends on differences, but any symbolic system weakens differences [25]. Third, the so-called "alignment problem" thus becomes incomplete because human values themselves contain a large number of contextual and informal factors that cannot be fully encoded into a model.

In consciousness theory, negative ontology further deepens the so-called "difficult problem." David Chalmers points out that why physical processes produce subjective experiences is the most central and intractable problem in consciousness research [31]. Negative ontology offers a more radical explanatory path: this problem is "difficult" not because we haven't found the correct scientific theory, but because it is structurally irreducible. In other words, any third-person description of the brain can only cover a portion of consciousness, and cannot exhaust the first-person experience itself.

This view implies that subjective experience possesses an "inherent untranslatability." We can describe neural activity, information processing, and even behavioural patterns, but we cannot fully translate the "feeling of pain" or the "experience of colour" into physical language. This irreducibility can be compared to the "incompressibility" in information theory: certain information cannot be replaced by a more concise description. Furthermore, this makes the question of "whether AI possesses consciousness" resemble an undecidable proposition: Even if a system behaves completely identically to a human, we still cannot determine whether it truly possesses subjective experience. This dilemma is logically analogous to the incompleteness revealed by Kurt Gödel [32].

When this line of thought is extended to the question of "whether reality is computable," its impact becomes even more fundamental. An important position in modern science holds that the universe is inherently a computable structure, an idea developed in digital physics and information ontology. However, negative ontology poses a direct challenge to this: if reality itself contains an unformatable dimension, then any computational model can only capture a part of it, not the whole. In other words, reality is not equivalent to a computable structure, but contains an uncomputable "remainder."

This conclusion can be seen as an ontological generalization of Kurt Gödel's incompleteness theorems: not only can formal systems not be self-complete, but reality itself cannot be completely described by any single formal system [32]. It also echoes Alan Turing's discoveries about uncomputability, such as the computational boundaries revealed by the halting problem [30]. If similar structural constraints exist in reality, it means that certain processes cannot be simulated by algorithms in principle. Thus, the idea of a "simulated universe" is weakened, because simulation only reproduces structure, not existence itself.

In summary, these areas can be unified into a three-layered structure: the top layer is the representable layer, including AI models, scientific theories, and mathematical structures; the middle layer is existence itself; and the bottom layer is the unrepresentable negative dimension. The core claim of the denial of ontology is that existence always transcends its representation, and any model or theory can only partially approximate reality.

Therefore, we present the following Conclusion 1 regarding the Potential Impact of Negative Ontology:

**Conclusion 1 (Potential Impact of Negative Ontology)**: *the world is not a problem that can be fully described and calculated, but rather a structure that inherently contains indescribability and inexhaustibility. This also means that "incomprehensibility" is not merely a limitation of human cognitive abilities, but is a fundamental characteristic of existence itself.*

**Conclusion 1 reveals that $N(x)$ is not merely an epistemic limit but a source of potential, which forms the philosophical guidance for inquiry and practice.**

## 3. The Expansion Principle of Negative Ontology

Building on Definition 1 (Negative Ontology) and Conclusion 1 (Potential Impact of Negative Ontology), in this section, we aim to translate these philosophical foundations into a formal, operationalizable framework. Accordingly, we propose Principle 1: the Expansion Principle of Negative Ontology.

**Principle 1 (Expansion Principle of Negative Ontology):** *For any research domain* $X$*, approaches under traditional ontology define it through observable, measurable, or codifiable elements* $D(X)$*. Negative Ontology introduces previously excluded or undefined dimensions* $N(X)$ *(gaps, absences, contradictions, affective forces, or undefinable truths) producing an expanded domain of inquiry:*

$$X^{+} = D(X) \cup N(X),\ X^{+} \supset X.$$

The key insight for this principle is that "absence," "nonexistence," or "undefinability" is generative: it produces new interpretive, methodological, and practical possibilities rather than representing a deficiency.

The remainder of this section presents the theoretical validity of Principle 1, its operationalization through the instantiated three expansion mechanisms, and the integrated impact of the three mechanisms for expansion of possibility.

## 3.1 Theoretical validity

The validity of Principle 1 (Expansion Principle of Negative Ontology) follows by the following Axioms 1-3 and Propositions 1-3:

**Axiom 1 (Ontological Incompleteness)**: *The existence of any object exceeds its describable set:* $D(x) \subsetneq x$*.*

**Axiom 2 (Constitutive Negativity)**: *The negative dimension* $N(x) \neq \emptyset$*, and is necessary for the generation of meaning/function.*

**Axiom 3 (Open System)**: *Any research system* $S$ *can be written as* $S = S_D \cup S_N$*, where* $S_N$ *cannot be fully formalized or closed.*

**Proposition 1 (Domain Expansion)**: *Including N(x) expands the research domain.*

**Proposition 2 (Generative Enhancement)**: *The negative dimension enhances the system's generative capacity.*

**Proposition 3 (Open-System Superiority)**: *Open systems are superior to closed systems (in complex environments).*

With respect to Definition 1 (Negative Ontology) and Conclusion 1 (Potential Impact of Negative Ontology), these axioms can be seen as structurally isomorphic to Martin Heidegger's "manifestation–concealment structure," [24] Jacques Derrida's "différance" [25] and Lao Tzu's "Tao," [26] and Kurt Gödel's incompleteness theorems [32], corresponding respectively to the ontological, linguistic/meaning-based, and formal system levels.

Building on the three foundational axioms of Negative Ontology, Ontological

Incompleteness, Constitutive Negativity, and Open System, we can rigorously assess the validity of the corresponding propositions regarding the expansion and generative capacity of research and practice.

Proposition 1 follows directly from the axioms of Ontological Incompleteness and Constitutive Negativity. Axiom 1 establishes that the describable component of any object, $D(x)$, is strictly a subset of the object's total existence, $x(D(x) \subsetneq x)$. Consequently, a research program that considers only $D(x)$ necessarily neglects aspects of the object that cannot be fully described. Axiom 2 further stipulates that the negative dimension $N(x)$ is non-empty and constitutive of meaning or function. When $N(x)$ is incorporated into the scope of study, the research domain expands from the limited, fully describable subset $D(x)$ to the union $D(x) \cup N(x)$. In other words, integrating the negative dimension enlarges both the conceptual and practical boundaries of inquiry, allowing scholars to account for dimensions of objects previously excluded from systematic analysis. Therefore, Proposition 1 is a direct consequence of the axioms and holds as a logically necessary claim.

Proposition 2 builds explicitly on Axiom 2, which frames $N(x)$ as essential for the generation of meaning and function. In complex systems, generative capacity refers to the system's ability to produce new interpretations, behaviours, or outcomes. By including the negative dimension, the system no longer operates solely within the constraints of $D(x)$; it also leverages the potentiality embedded in $N(x)$. This allows for emergent dynamics that are unavailable in a strictly positive, fully describable system. Formally, if we denote the system's generative potential as $Gen(S)$, then $Gen(S_D \cup S_N) > Gen(S_D)$. Thus, the negative dimension functions not as an extraneous or residual component but as an essential source of novelty and meaning. Proposition 2 is therefore a natural extension of the axioms: the inclusion of $N(x)$ enhances the system's ability to generate new states, interpretations, and functions.

Proposition 3 derives from Axiom 3, which formalizes the research system as $S = S_D \cup S_N$, with $S_N$ being inherently non-formalizable and resistant to closure. In complex, dynamic environments characterized by uncertainty and nonlinearity, closed systems that operate solely on $S_D$ are limited to predictable and fully described variables, making them brittle and prone to failure. Open systems, by contrast, integrate $S_N$ and thus remain sensitive and responsive to emergent phenomena. The open configuration allows the system to exploit the generative capacity of the negative dimension, producing adaptive responses to unanticipated conditions. Empirical and theoretical evidence from complex systems theory supports this view: systems that accommodate incompleteness, ambiguity, and relational emergence outperform rigid, closed architectures in terms of flexibility, robustness, and adaptive performance. Consequently, Proposition 3 is well justified: in complex environments, open systems are superior to closed ones.

The three axioms provide the ontological and methodological foundation for the Expansion Principle. Axiom 1 (Ontological Incompleteness) establishes that the describable component $D(x)$ is always partial ($D(x) \subsetneq x$), implying that any traditional, positivist approach inherently underrepresents the full domain. Axiom 2 (Constitutive Negativity) asserts that the negative dimension $N(x) \neq \emptyset$ is not incidental but necessary for the generation of meaning and function. Axiom 3 (Open System) formalizes research systems as $S = S_D \cup S_N$, where $S_N$ is irreducible and cannot be fully formalized, ensuring the system's openness to emergent possibilities.

The corresponding propositions translate these axioms into actionable insights. Proposition 1 (Domain Expansion) shows that including $N(x)$ systematically enlarges the research domain, providing a formal rationale for expanding beyond observable or measurable elements. Proposition 2 (Generative Enhancement) highlights that the negative dimension is generative, producing novel meanings, interpretations, and functionalities. Proposition 3 (Open-System Superiority) demonstrates that open systems leveraging $N(x)$ are better suited to complex and dynamic environments than closed, purely positive systems.

Together, these axioms and propositions directly support the Expansion Principle: they justify that research should not be confined to $D(X)$ alone but should incorporate $N(X)$, producing an expanded domain $X^+ = D(X) \cup N(X)$. This formalizes the core insight of Negative Ontology: absence, incompleteness, and undefinability are not deficiencies to be eliminated but generative forces that broaden theoretical, methodological, and practical horizons. The theoretical validity for the Expansion Principle of Negative Ontology is summarized in Table 1.

## 3.2 Expansion mechanisms

The Expansion Principle of Negative Ontology can be operationalized through three complementary mechanisms: Boundary Relaxation, Semantic Reversal, and Structural Reframing. Each of these mechanisms directly instantiates one of the corresponding propositions (Proposition 1 (Domain Expansion), Proposition 2 (Generative Enhancement), and Proposition 3 (Open-System Superiority)) into actionable, operational practices.

**Expansion Mechanism 1 (Boundary Relaxation / Domain Expansion):** *Traditional research achieves precision by strictly defining its objects of study, but this also narrows the scope of inquiry. Introducing $N(x)$ implies:*

- *Expanding the variable space,*
- *Incorporating unstructured or previously excluded information,*
- *Redefining what counts as "relevant".*

**Expansion Mechanism 2 (Semantic Reversal / Generative Enhancement):** *In conventional paradigms:*

- *Uncertainty = problem,*
- *Absence = error.*

*Under the Expansion Principle:*

- *Uncertainty = a generative condition,*
- *Absence = a source of information.*

*This reversal transforms what was previously excluded into the core of research focus.*

**Expansion Mechanism 3 (Structural Reframing / Open-System Superiority):** *The system shifts from a closed form:*

$$f: X \to Y$$

*to an open form:*

$$f: (X \cup N_X) \to (Y \cup N_Y)$$

*As a result, the system transforms from a static structure into a dynamic process, capable of generating new states and interpretations.*

**Table 1. Theoretical validity for the Expansion Principle of Negative Ontology**

| Component | Description | Correspondence / Effect |
|---|---|---|
| **Axiom 1: Ontological Incompleteness** | The existence of any object exceeds its fully describable set: $D(x) \subsetneq x$. | Establishes that research based only on $D(x)$ is inherently partial; motivates the inclusion of negative dimensions. |
| **Axiom 2: Constitutive Negativity** | The negative dimension $N(x)$ is necessary for generating meaning and function. | Provides the generative basis for novel interpretations, emergent phenomena, and functional capacities. |
| **Axiom 3: Open System** | Any research system ($S = S_D \cup S_N$), where $S_N$ cannot be fully formalized or closed. | Ensures adaptability and responsiveness in complex environments; allows $N(x)$ to operate within systems. |
| **Proposition 1: Domain Expansion** | Including $N(x)$ expands the research domain. | Extends the scope of inquiry from $D(x)$ to $D(x) \cup N(x)$; foundational for the Expansion Principle. |
| **Proposition 2: Generative Enhancement** | The negative dimension enhances the system's generative capacity. | Demonstrates that $N(x)$ produces new meanings, functions, or emergent outcomes. |
| **Proposition 3: Open-System Superiority** | Open systems are superior to closed systems in complex environments. | Shows that $S = S_D \cup S_N$ is more robust and adaptive than $S_D$-only systems. |
| **Principle 1: Expansion Principle of Negative Ontology** | For any research domain $X$ incorporating negative dimensions $N(X)$ expands the domain: $X^+ = D(X) \cup N(X)$. | Synthesizes Axioms 1-3 and Propositions 1-3; formalizes how negative ontology systematically enlarges the scope of research and practice. |

Expansion Mechanism 1 (Boundary Relaxation / Domain Expansion) addresses the limitations of traditional research that relies on strictly defined objects to ensure precision. By incorporating $N(x)$, researchers expand the variable space, include previously unstructured or excluded information, and redefine relevance, thereby enlarging the domain of inquiry beyond the constraints of fully describable elements.

Expansion Mechanism 2 (Generative Enhancement / Semantic Reversal) reconceptualizes what is traditionally considered problematic or deficient: uncertainty and absence are no longer treated as obstacles but as generative conditions and sources of information. This inversion transforms previously marginalized dimensions into central analytic and practical foci, enabling research to exploit the productive potential of gaps, contradictions, and indeterminacies.

Expansion Mechanism 3 (Structural Reframing / Open-System Superiority) formalizes this shift by moving systems from a closed mapping $f: X \to Y$ to an open mapping $f: (X \cup N_X) \to (Y \cup N_Y)$, turning static structures into dynamic processes capable of producing new states, interpretations, and functional outcomes.

Together, these expansion mechanisms operationally instantiate the Expansion Principle of Negative Ontology: they show that the inclusion of negative dimensions does not merely add complexity but systematically generates new interpretive, methodological, and practical possibilities, demonstrating that absence, incompleteness, and undefinability are intrinsically productive rather than deficient.

### 3.3 Integrated impact: expansion of possibility

By synergistically applying these three mechanisms, the framework operationalizes the Expansion Principle in a way that:

- Broadens ontological horizons by including previously excluded dimensions.
- Enhances generative capacity by leveraging uncertainty and absence to create new knowledge, models, and practices.
- Improves adaptability and resilience of systems and methodologies by adopting open-system thinking.
- Provides greater theoretical insight and practical value by systematically incorporating what is inherently indescribable, incomplete, or emergent.

In essence, the three mechanisms collectively ensure that Negative Ontology is not merely a philosophical position but a practical, operationalizable framework, capable of expanding the scope of research and practice.

## 4. Negative Ontology Reframing Methodology

Grounded in the three expansion mechanisms under the Expansion Principle of Negative Ontology, we propose the Negative Ontology Reframing Methodology, which is a systematic approach established to reframe the scope of research and practice by deliberately engaging with both what is known and what is absent or indescribable. The methodology unfolds in six interconnected steps:

**Step 1 (Define the Domain and Object)**: *The process begins by clearly specifying the focal object, phenomenon, problem, theory, or system ($X$), along with its boundaries and underlying assumptions. This step establishes the scope and context for analysis, making explicit what is fully describable ($D(X)$) while acknowledging the limitations of conventional, positive ontology. By doing so, it prepares the ground for incorporating elements that are inherently absent, ambiguous, or otherwise excluded.*

**Step 2 (Surface the Negative Dimension ($\boldsymbol{N}(\boldsymbol{X})$))**: *Next, the methodology actively identifies gaps, absences, unknowns, and unobserved elements, which are aspects that are difficult or impossible to describe within conventional frameworks. This step, driven by Expansion Mechanism 1 (Boundary Relaxation / Domain Expansion), expands the conceptual space by including previously ignored or invisible dimensions. It emphasizes that absence and incompleteness are not merely limitations but generative forces that shape the domain itself.*

**Step 3 (Integrate $\boldsymbol{N}(\boldsymbol{X})$ (Reframe))**: *Having surfaced these negative dimensions, the*

*methodology explicitly incorporates them into the analysis. The problem or system is reformulated as* $X^{+} = D(X) \cup N(X)$, *integrating both the describable and the indescribable. Using mechanisms such as Expansion Mechanism 1 (Boundary Relaxation / Domian Expansion) and Expansion Mechanism 2 (Semantic Reversal / Generative Enhancement), previously marginal or excluded elements become productive, creating tension between known and unknown dimensions. This tension becomes a source of creativity, transforming gaps into essential components of theory and practice.*

**Step 4 (Activate Expansion Mechanisms)**: *At this stage, the methodology operationalizes the Expansion Principle by systematically applying its core mechanisms. This includes leveraging the tension between* $D(X)$ *and* $N(X)$ *to stimulate the generation of new concepts, maintaining openness to* $N(X)$ *for adaptability, and continually expanding the domain of inquiry. Through Expansion Mechanism 1 (Boundary Relaxation / Domain Expansion), Expansion Mechanism 2 (Semantic Reversal / Generative Enhancement), and Expansion Mechanism 3 (Structural Reframing / Open-System Superiority), this step ensures that the system is generative, expansive, and resilient.*

**Step 5 (Generate and Explore Emergences)**: *With the framework in place, new hypotheses, strategies, models, and practices are produced, following Expansion Mechanism 2 (Semantic Reversal / Generative Enhancement). This step exploits the interplay between positive and negative dimensions to foster novelty, creativity, and interpretive richness. Emergent phenomena arise as the interaction between what is known and what is absent generates new insights and possibilities, demonstrating that the unknown is not a void but a productive space for discovery.*

**Step 6: (Evaluate and Iterate (Open Loop))**: *Finally, the methodology emphasizes continuous assessment and iterative refinement, following Expansion Mechanism 3 (Structural Reframing / Open-System Superiority). Outcomes are evaluated using expanded criteria, and engagement with* $N(X)$ *is deepened to maintain adaptability. By treating the system as an open, evolving loop, the methodology ensures robustness, reflexivity, and systemic resilience, highlighting the open-ended nature of cognition and reinforcing the central insight that negative dimensions are constitutive, not peripheral, to understanding.*

A logical visualization of the six steps in the Negative Ontology Reframing Methodology is presented in Fig. 1, with the corresponding key points summarized in Table 2.

## 5. Application of the Negative Ontology Reframing Methodology Across Disciplines

Based on the Negative Ontology Reframing Methodology, established under the Expansion Principle of Negative Ontology, we revisited three bodies of work that individually explore the reframing of research and practice in the disciplines of Leadership [15], Information Systems (IS) [16], and Machine Learning (ML) [17]. Each discipline is rebuilt with the Negative Ontology Reframing Methodology to illustrate how the six-step methodology generally operationalizes Negative Ontology to expand theoretical, methodological, and practical horizons across disciplines.

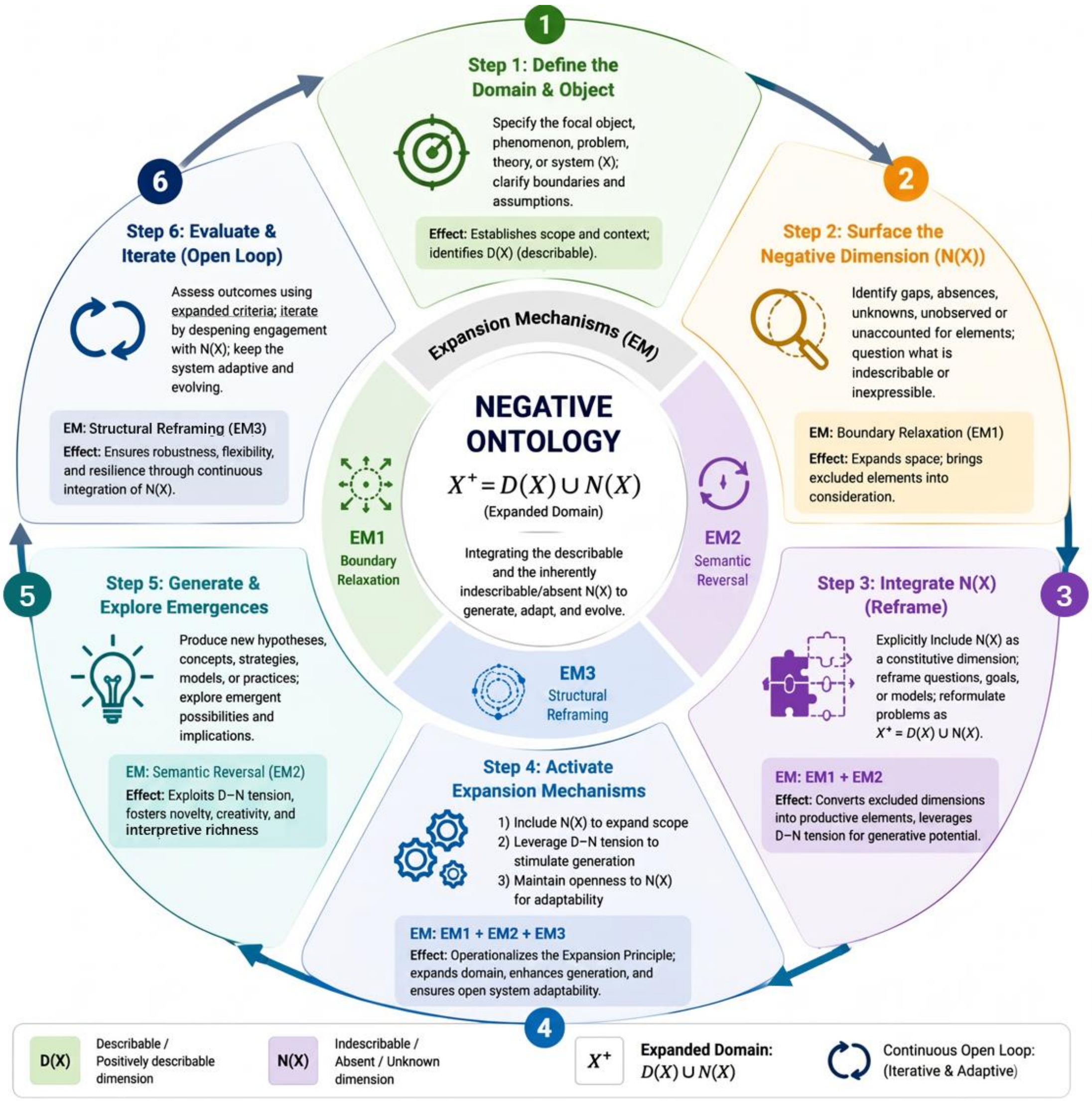

**Figure 2. logical visualization of the six steps in the Negative Ontology Reframing Methodology.**

**Acknowledgement:** This image was generated by ChatGPT with provided structural information and was manually refined by the authors to accurately illustrate the contents.

**Table 2. Negative Ontology Reframing Methodology: Steps mapped to expansion mechanism(s) and corresponding effect and implication**

| Step | Description | Corresponding Expansion Mechanism(s) | Effect / Role | Negative Ontology Implication |
|---|---|---|---|---|
| **Step 1: Define the Domain & Object** | Specify the focal object, phenomenon, problem, theory, or system ( $X$ ); clarify boundaries and assumptions. | None explicitly (identifies $D(X)$) | Establishes scope and context for reframing; identifies what is fully describable ($D(X)$). | Recognizes the limits of positive ontology and sets the stage for including inherently indescribable or absent dimensions ( $N(X)$ ) in research and practice. |
| **Step 2: Surface the Negative Dimension ($N(X)$)** | Identify gaps, absences, unknowns, unobserved or unaccounted-for elements; question what is indescribable or inexpressible. | **Boundary Relaxation / Domain Expansion (EM1)** | Expands the conceptual/variable space; brings previously excluded elements into consideration. | Highlights the constitutive role of absence and incompleteness in shaping the domain, revealing that what is excluded or unknown is generative. |
| **Step 3: Integrate $N(X)$ (Reframe)** | Explicitly include $N(X)$ as a constitutive dimension; reframe questions, goals, or models; reformulate problems as $X^+ = D(X) \cup N(X)$. | **EM1 + Semantic Reversal / Generative Enhancement (EM2)** | Converts previously excluded or marginal dimensions into productive elements; leverages tension between $D(X)$ and $N(X)$ for generative potential. | Treats absence, ambiguity, and indefinability as constitutive for knowledge generation, transforming gaps into essential components of theory and practice. |
| **Step 4: Activate Expansion Mechanisms** | Apply the mechanisms systematically: 1) include $N(X)$ to expand scope, 2) leverage D–N tension ($D(X) \cup N(X)$) to stimulate generation, 3) maintain openness to $N(X)$ for adaptability. | **EM1 + EM2 + Structural Reframing / Open-System Superiority (EM3)** | Operationalizes the Expansion Principle; simultaneously expands domain, enhances generation, and ensures open-system adaptability. | Embeds the Expansion Principle in practice, demonstrating that engagement with negative dimensions drives both conceptual and functional expansion. |
| **Step 5: Generate & Explore Emergences** | Produce new hypotheses, concepts, strategies, models, or practices; explore emergent possibilities and implications. | **EM2** | Exploits the productive tension between positive and negative dimensions; fosters novelty, creativity, and interpretive richness. | Reveals that the interaction between known and unknown dimensions generates novel insights, solutions, and emergent phenomena across domains. |
| **Step 6: Evaluate & Iterate (Open Loop)** | Assess outcomes using expanded criteria; iterate by deepening engagement with $N(X)$; keep the system adaptive and evolving. | **EM3** | Ensures robustness, flexibility, and resilience; continuously integrates $N(X)$ to maintain open, adaptive research or practice systems. | Demonstrates that continuous engagement with the negative dimension sustains adaptability, reflexivity, and systemic resilience, highlighting the open-ended nature of cognition. |

## 5.1 Leadership

Applying the six-step Negative Ontology Reframing Methodology [15]:

1) **Define the Domain & Object** – Traditional approaches often treat leadership as a set of traits, behaviors, or outcomes that can be objectively measured.
2) **Surface the Negative Dimension (N(X))** – Ambiguities, ideological absence, and "floating" signifiers of leadership are revealed.
3) **Integrate (N(X)) (Reframe)** – These negative dimensions are treated as constitutive, making leadership a generative site of meaning rather than a fixed entity.
4) **Activate Expansion Mechanisms** – Boundary relaxation and semantic reversal reposition uncertainty and absence as generative, structural reframing ensures a dynamic interpretation of leadership.
5) **Generate & Explore Emergences** – Multiple interpretations, organizational myths, and emergent power dynamics are examined.
6) **Evaluate & Iterate (Open Loop)** – Reflexive evaluation of leadership practices emphasizes ideological influence and the interplay between formal authority and emergent social meaning.

In summary, leadership research shifts from measuring traits to understanding leadership as an emergent, ideologically productive phenomenon, broadening both theoretical insight and practical interventions. More details are provided in Table 3.

## 5.2 Information Systems (IS)

Through the reframing methodology under Negative Ontology [16]:

1) **Define the Domain & Object** – The focus extends beyond formal social/material structures to the relational, affective engagement between users and technology.
2) **Surface the Negative Dimension (N(X))** – Gaps, inconsistencies, unarticulated motivations, and preconscious emotional responses are identified.
3) **Integrate (N(X)) (Reframe)** – These affective and unstructured dimensions are incorporated as central to understanding system enactment.
4) **Activate Expansion Mechanisms** – Boundary relaxation expands the conceptual space to include affect; semantic reversal treats uncertainty as a productive influence; structural reframing allows for dynamic co-evolution of social and technological practices.
5) **Generate & Explore Emergences** – New patterns of relational agency, system use, and emergent technological adaptations are explored.
6) **Evaluate & Iterate (Open Loop)** – Reflexive assessment emphasizes ethical sensitivity, affective responses, and the continuous interplay between users and systems.

In summary, IS research evolves from deterministic models to a richer, affectively and relationally mediated understanding of human-technology interaction. More details are provided in Table 4.

## 5.3 Machine Learning (ML)

Applying the methodology under Negative Ontology [17]:

1) **Define the Domain & Object** – Standard paradigms assume the existence of an

objective “true target” (TT) for supervision.

2) **Surface the Negative Dimension (N(X))** – Explicitly positing the opposite assumption that the TT does not objectively in the real world, multiple inaccurate true targets (MIATTs) are defined, highlighting ambiguity and uncertainty for democratic supervision.
3) **Integrate (N(X)) (Reframe)** – MIATTs are incorporated into evaluation and learning procedures, moving from an authoritative ground truth to democratic/pluralistic supervision for ML.
4) **Activate Expansion Mechanisms** – Boundary relaxation expands the supervision space from authoritative to democratic; semantic reversal treats uncertainty and conflicts as productive signals; structural reframing ensures that the ML system remains adaptive to multiple / democratic perspectives.
5) **Generate & Explore Emergences** – ML reconciles conflicting perspectives, producing robust, human-centered outputs.
6) **Evaluate & Iterate (Open Loop)** – Continuous integration of new perspectives, feedback loops, and pluralistic evaluation metrics enhance adaptability and interpretability.

In summary, ML research moves from approximating an assumed objective target to continuously exploring cognition boundaries under inherently uncertainty, embracing participatory and pluralistic (democratic) supervision. More details are provided in Table 5.

### 5.4 Summary Across Disciplines

Applying the Negative Ontology Reframing Methodology consistently demonstrates that:

- **Boundary Relaxation** enlarges the conceptual and operational space in each domain.
- **Semantic Reversal** converts uncertainty, absence, or contradiction into productive sources of insight.
- **Structural Reframing** ensures adaptability, robustness, and emergent functionality.

Across Leadership, IS, and ML, the methodology generally operationalizes the Expansion Principle: previously excluded or undefined dimensions $N(X)$ are not limitations but generative elements that expand theoretical, methodological, and practical possibilities.

## 6. Discussion of Negative Ontology as an Expansion Principle

The formal definition of Negative Ontology (Definition 1) and the formal conclusion regarding its potential impact (Conclusion 1) jointly establish the ontology of non-existence for this study. To translate these foundations into a practical, operationalizable framework, we propose Principle 1: the **Expansion Principle of Negative Ontology** ($X^+ = D(X) \cup N(X)$). Principle 1 is operationalized through three complementary **Expansion Mechanisms**: **Boundary Relaxation, Semantic Reversal, and Structural Reframing**. They together enable broader ontological horizons, enhanced generative capacity, improved adaptability and resilience, and greater theoretical and practical significance.

Table 3. Negative Ontology Expansion Principle-Informed Reframing of Leadership (Six-Step Framework)

| Step | Leadership Reframing | Corresponding Expansion Mechanism(s) | Effect / Role | Negative Ontology Implication |
|---|---|---|---|---|
| **Step 1: Define the Domain & Object** | Leadership is initially defined as a focal object in organizational studies: roles, traits, behaviors, influence structures, or decision-making authority ($X$). Traditional leadership research assumes leadership as a positive, observable entity. | None explicitly (identifies $D(X)$) | Establishes the baseline ontology of leadership as something that can be identified, measured, or located in individuals or systems. | It reveals that this "positive object" is already a constructed assumption: leadership appears as something stable only within an epistemic framing that ignores absence, ambiguity, and ideological construction. |
| **Step 2: Surface the Negative Dimension ($N(X)$)** | Identify what is missing or ungraspable in leadership: ambiguity of authority, emotional projection, symbolic authority, ideological desire, and interpretive instability. Leadership is also defined by what it fails to concretely be. | **Boundary Relaxation / Domain Expansion (EM1)** | Expands leadership beyond observable traits into symbolic, affective, and ideological space. | Leadership is revealed as structurally incomplete: it is always partially absent, always exceeding empirical description. What appears as leadership depends on what is not explicitly present but still felt or assumed. |
| **Step 3: Integrate ($N(X)$) (Reframe)** | Leadership becomes redefined as an "empty signifier" which is structured by absence and overdetermination of meaning. | **EM1 + Semantic Reversal / Generative Enhancement (EM2)** | Converts absence into constitutive structure; leadership becomes meaningful precisely through indeterminacy and openness. | Leadership is no longer something that exists, but something that emerges through lack. Its identity is fundamentally negative: it exists as an interpretive vacancy filled by discourse, expectation, and desire. |
| **Step 4: Activate Expansion Mechanisms** | Leadership is analyzed as an open system where meanings continuously shift across discourse, myth, and organizational practice. It cannot be stabilized as a fixed entity. | **EM1 + EM2 + Structural Reframing / Open-System Superiority (EM3)** | Leadership is reframed from a bounded object to a dynamic semiotic field producing multiple meanings. | Leadership becomes an open ideological system, not a measurable variable. It is structurally dependent on its own instability and incompleteness to function socially. |
| **Step 5: Generate & Explore Emergences** | Investigate emergent leadership phenomena: charismatic attribution, follower projection, organizational myths, symbolic authority formation, and shifting legitimacy structures. | **EM2** | Absence and ambiguity become productive forces generating meaning, authority, and organizational coherence. | Leadership is revealed as a productive absence: it generates coordination, identity, and belief not through essence but through interpretive surplus and symbolic construction. |
| **Step 6: Evaluate & Iterate (Open Loop)** | Leadership analysis shifts from measuring traits/outcomes to examining how leadership discourse, myths, and narratives continuously reproduce authority and desire. | **EM3** | Leadership becomes an evolving ideological process rather than a stable construct. | Leadership is never fully captured; it is continuously reconstructed through discourse and practice. Research and practice focus on how leadership is produced, not what it is. |

**Table 4. Negative Ontology Expansion Principle-Informed Reframing of Information System (Six-Step Framework)**

| Step | Information System (IS) Reframing | Corresponding Expansion Mechanism(s) | Effect / Role | Negative Ontology Implication |
|---|---|---|---|---|
| **Step 1: Define the Domain & Object** | IS is traditionally defined as the interaction between social structures (rules, norms, meanings) and material structures (technologies, infrastructure) ( $X$ ), focusing mainly on rational/judgmental agency and observable system use. | None explicitly (identifies $D(X)$) | Establishes a bounded ontology of IS as a system of describable social/material elements and rational actions. | It reveals that this ontology is already partial: it excludes affect, absence, and preconscious experience. IS is initially framed as overly "positive," assuming completeness where there is structural incompleteness. |
| **Step 2: Surface the Negative Dimension ($N(X)$)** | Identify what IS research traditionally ignores: affective states, preconscious drives, emotional tensions, ambiguity in technology use, and gaps between structure and practice. | **Boundary Relaxation / Domain Expansion (EM1)** | Expands IS beyond formal systems into lived, affective, and experiential dimensions. | IS is not fully explainable through social/material structures. Instead, it is structured by incompleteness (gaps) that shape how technology is experienced but not fully articulated. |
| **Step 3: Integrate ($N(X)$) (Reframe)** | IS becomes a Gap–Affect–Practice system: gaps in structure generate affective responses, which lead to identification processes and emergent practices. | **EM1 + Semantic Reversal / Generative Enhancement (EM2)** | Converts absence (gaps, uncertainty, emotional ambiguity) into a central generative driver of IS dynamics. | The core of IS is not systems or technologies, but negativity experienced as affect. Agency emerges not from structure but from how subjects emotionally interpret structural absence. |
| **Step 4: Activate Expansion Mechanisms** | IS is reframed as an open sociomaterial-affective system, where structures, gaps, and affects continuously co-produce practices and realities. | **EM1 + EM2 + Structural Reframing / Open-System Superiority (EM3)** | Transforms IS from a closed system of interaction into an open, evolving process of co-constitution between humans, technology, and affect. | IS becomes fundamentally non-closed and non-stable: meaning, agency, and structure continuously emerge through interaction with negativity and affective experience. |
| **Step 5: Generate & Explore Emergences** | Emergent IS phenomena include: affect-driven technology use, identity formation through emotional engagement, and practice evolution triggered by structural gaps. | **EM2** | Exploits tension between structure and absence to generate new practices, behaviors, and interpretations. | IS is not a system of optimization or efficiency, but a system of emergent meaning production through affectively experienced incompleteness. |
| **Step 6: Evaluate & Iterate (Open Loop)** | IS evaluation shifts from efficiency and rational outcomes to analyzing affective experience, relational agency, ethical engagement, and iterative practice formation. | **EM3** | Ensures IS remains adaptive, reflective, and continuously evolving through engagement with gaps and affect. | IS knowledge is never complete; it is continuously reconstructed through experience, emotion, and structural absence. Research and practice become an ongoing engagement with incompleteness rather than closure. |

**Table 5. Negative Ontology Expansion Principle-Informed Reframing of Machine Learning (Six-Step Framework)**

| Step | Machine Learning (ML) Reframing | Corresponding Expansion Mechanism(s) | Effect / Role | Negative Ontology Implication |
|---|---|---|---|---|
| **Step 1: Define the Domain & Object** | Traditional ML defines the domain as learning a mapping from a to the True Target (TT) ($X$), assumed to exist objectively in the real world. This includes supervised, weakly supervised, and noisy-label learning paradigms. | None explicitly (identifies $D(X)$) | Establishes the baseline assumption that ML is about approximating an objective ground truth through labeled data. | It reveals that this assumption is ontologically loaded: it presupposes stability, objectivity, and definability of truth. This sets up the target for deconstructions. (Existence assumption of TT) |
| **Step 2: Surface the Negative Dimension ($N(X)$)** | Identify what ML ignores: ambiguity, disagreement between annotators, cultural/contextual variation, epistemic uncertainty, and the non-existence of a single TT. | **Boundary Relaxation / Domain Expansion (EM1)** | Expands ML space beyond labeled datasets into plural, inconsistent, and socially constructed signals. | Truth is not missing or noisy but structurally absent or undefinable. This introduces the idea that supervision itself contains irreducible plurality. (Non-existence assumption of TT) |
| **Step 3: Integrate ($N(X)$) (Reframe)** | Replace TT with a democratic/plural structure: multiple inconsistent annotations, perspectives, and interpretations (MIATTs (Multiple Inaccurate True Targets)). | **EM1 + Semantic Reversal / Generative Enhancement (EM2)** | Converts disagreement and inconsistency into core learning material rather than noise. | There is no authoritative ground truth. Instead, learning operates over distributed, partial, and conflicting realities, where "truth" is an emergent construct rather than an existing object. |
| **Step 4: Activate Expansion Mechanisms** | ML becomes a system of Democratic Supervision, where experts, non-experts, and machines jointly generate supervision signals; learning is based on aggregated plural perspectives. | **EM1 + EM2 + Structural Reframing / Open-System Superiority (EM3)** | Transforms ML from a closed optimization toward TT into an open system integrating multiple epistemic sources. | ML is no longer a truth-approximation system but an open epistemic ecology where knowledge emerges from democratically structured disagreement and participation. |
| **Step 5: Generate & Explore Emergences** | Models reconcile MIATTs via frameworks like LAF (Logical Assessment Formula) and UTTL (Undefinable True Target Learning), producing outputs that reflect collective interpretation boundaries. | **EM2** | Exploits tension between inconsistent targets to generate richer, more robust representations and predictions. | Learning becomes a process of constructing coherence from irreducible inconsistency, not discovering hidden truth. The model reveals cognitive boundaries rather than resolving them. |
| **Step 6: Evaluate & Iterate (Open Loop)** | Evaluation shifts from accuracy against TT to assessing consistency across MIATTs, interpretability under disagreement, and robustness across plural signals. Continuous iterative refinement integrates new perspectives. | **EM3** | Ensures adaptive, evolving ML systems that incorporate ongoing pluralism and uncertainty. | ML becomes an open-ended epistemic process, continuously reshaped by democratic participants, interpretations, and contradictions, never converging to a final truth state but continuously revealing cognition boundaries. |

Building on these mechanisms, the **Negative Ontology Reframing Methodology** is established as a systematic approach to expand the scope of research and practice by deliberately engaging with both what is known and what is absent or indescribable. Its application across the disciplines of Leadership, Information Systems, and Machine Learning demonstrates the methodology's generalizability and its capacity to extend inquiry beyond the limits of conventional, object-centered, or fully formalizable paradigms.

Collectively, these contributions position **Negative Ontology as an Expansion Principle** as a framework with substantial theoretical significance, methodological innovation, and the potential to challenge entrenched assumptions of computability, strong reductionism, and completeness in existing research and practice paradigms.

## 6.1 Theoretical significance

Negative Ontology as an Expansion Principle fundamentally shifts how we conceive of research and practice domains. Traditionally, inquiry has focused on studying well-defined objects, measurable phenomena, or fully describable entities. By emphasizing negative ontology, the framework redirects attention **from "object-centered research" to "boundary-centered research"**, highlighting the significance of gaps, absences, and incompleteness as constitutive elements of any system or phenomenon. This shift encourages scholars to investigate not only what is present and observable but also what is **structurally absent or excluded**, recognizing these elements as generative.

Additionally, the framework transforms the epistemic orientation of science itself. Conventional approaches often rely on the assumption that systems can be fully captured and modeled as deterministic entities. Negative Ontology reframing advances a perspective **from deterministic science to open-system science**, where incompleteness, unpredictability, and relational dynamics are central features of the research object. Such an approach aligns with contemporary complexity theory and systems thinking, highlighting the dynamic interplay between observable and unobservable dimensions.

## 6.2 Methodological innovation

Methodologically, Negative Ontology as an Expansion Principle operationalizes negative ontology by **transforming "unresearchable objects" into researchable ones**. Dimensions of a system or phenomenon that were traditionally considered inaccessible, undefined, or inexpressible ($N(X)$) are systematically incorporated into research design and practice analysis through the six-step methodology. By applying mechanisms such as **boundary relaxation, semantic reversal, and structural reframing**, scholars can engage with absences, contradictions, and uncertainties in a productive manner, expanding both the scope and depth of inquiry. This allows methodological innovation across disciplines, including the generation of new hypotheses, emergent practices, and adaptive models in complex environments.

## 6.3 Challenges to existing paradigms

Negative Ontology as an Expansion Principle directly challenges several entrenched assumptions in conventional research and practice paradigms:

1) **Computationalism** – Traditional models assume that all phenomena can be fully formalized and computed. Negative Ontology emphasizes the inherent incompleteness and non-computability of many real-world dimensions, highlighting the limits of purely algorithmic or formal approaches.
2) **Strong Reductionism** – Conventional reductionist approaches attempt to explain complex phenomena entirely through their constituent parts. Negative Ontology demonstrates that emergent properties often arise from absences, relational gaps, or unobserved dimensions, requiring holistic and open-system approaches.
3) **Completeness Assumption** – Many frameworks implicitly assume that a system can be exhaustively described and understood. Negative Ontology asserts that incompleteness is constitutive, and that gaps, absences, and undefinable elements are **not deficiencies but generative resources**.

### 6.4 Summary

In sum, Negative Ontology as an Expansion Principle provides a comprehensive lens for expanding theoretical, methodological, and practical boundaries of research and practice. Theoretically, it shifts focus from objects to boundaries and from deterministic to open systems. Methodologically, it operationalizes previously inaccessible dimensions into researchable constructs. Finally, it challenges dominant paradigms by questioning assumptions of computability, reductionism, and completeness, offering a more expansive, adaptive, and generative framework for inquiry across disciplines. The overview for the contents in this section is visually summarized in Fig. 2.

## 7. Future Work under Negative Ontology as an Expansion Principle

Negative Ontology as an Expansion Principle opens multiple avenues for future research, both in hypothesis testing and in methodological and applied extensions.

### 7.1 Hypotheses

**H1 (Incorporating N(X) enhances system robustness)**: This hypothesis posits that explicitly integrating the negative or unobserved dimensions of a system increases its resilience and stability under uncertainty, complexity, or perturbations. Empirical validation could involve comparative studies of systems with and without $N(X)$ integration across simulated and real-world contexts.

**H2 (Anomaly-based modeling improves generalization capacity)**: By treating gaps, absences, or anomalies as productive rather than as errors, systems can be trained to account for edge cases or unexpected conditions. This hypothesis can be tested in domains such as machine learning, complex network analysis, and organizational behavior, examining whether anomaly-informed models generalize better across novel conditions.

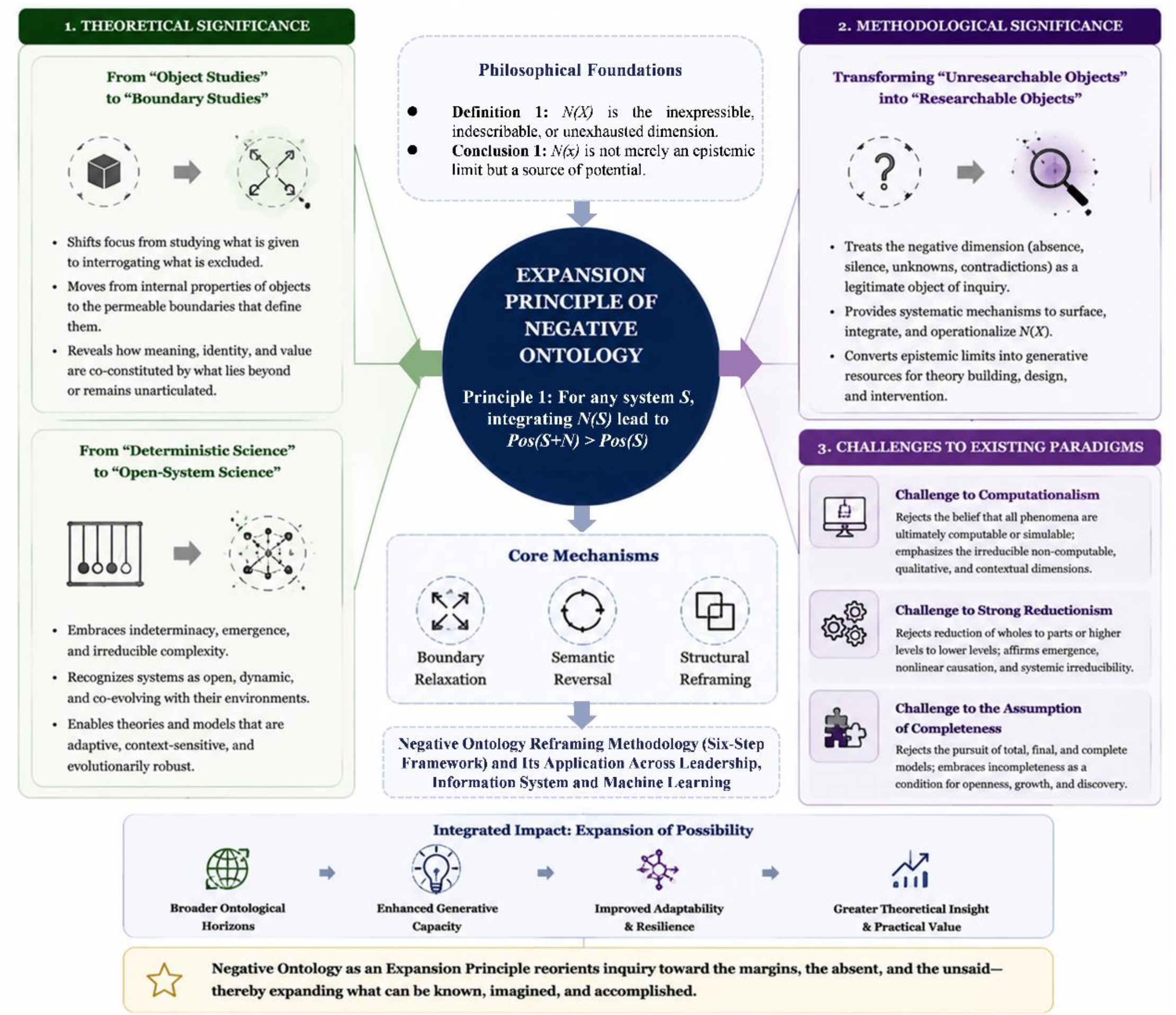

**Figure 2. Visualized overview for discussion of Negative Ontology as an Expansion Principle. Acknowledgement:** This image was generated by ChatGPT with provided structural information and was manually refined by the authors to accurately illustrate the contents.

**H3 (Open systems outperform closed systems in complex environments)**: Grounded in the open-system mechanism of Negative Ontology, this hypothesis predicts that systems designed to accommodate incompleteness, non-computability, or emergent dynamics will adapt more effectively than rigid, fully formalized systems when facing complex or dynamic environments.

## 7.2 Extensions

**Integration with Information Theory and Complexity**: Future work could link Negative Ontology to metrics of complexity, entropy, and information content, exploring how N(X) contributes to system-level information generation and interpretive richness. Such integration could formalize measures of generative potential arising from incompleteness or absence.

**Integration with AI Uncertainty Modelling**: Negative Ontology provides a natural foundation for uncertainty-aware AI and pluralistic supervision, moving beyond deterministic targets. Future research could explore frameworks for handling multiple, conflicting, or undefinable objectives (MIATTs), improving adaptability and robustness in AI systems.

**Development of "Negative Ontology-based AI"**: A promising applied direction is the construction of AI systems explicitly designed around negative ontology principles, where gaps, absences, and emergent possibilities are operationalized as productive inputs. Such systems could outperform conventional AI in tasks requiring adaptability, interpretive flexibility, or handling inherently incomplete information.

### 7.3 Summary

Future work under Negative Ontology as an Expansion Principle emphasizes empirical validation, integration with theoretical tools, and applied system development. By systematically exploring the effects of $N(X)$, anomaly modeling, and open-system architectures, researchers and practitioners can deepen understanding of generative inaccuracy, improve robustness and generalization, and create AI systems capable of operating effectively in complex, uncertain, or partially defined environments.

## 8. Conclusion

This article has presented Negative Ontology as an Expansion Principle as a novel ontological and philosophical foundation for research and practice. By systematically incorporating the "excluded or negative dimensions" ($N(X)$), the boundaries of inquiry are extended beyond what is fully describable, measurable, or formally codifiable. The Expansion Principle of Negative Ontology operationalizes this perspective through complementary expansion mechanisms (Boundary Relaxation, Semantic Reversal, and Structural Reframing), enabling researchers and practitioners to engage productively with gaps, absences, and uncertainties.

Importantly, the universality of this principle has been demonstrated across multiple domains, including Leadership, Information Systems, and Machine Learning, highlighting its adaptability and generalizability. Across these fields, the Negative Ontology Reframing Methodology reveals how engaging with what is traditionally excluded or unexpressed generates new theoretical insights, enhances generative capacity, and improves system adaptability and resilience.

Ultimately, Negative Ontology as an Expansion Principle challenges conventional assumptions that the limits of research and practice are determined largely by what is already known. Instead, it establishes that **the boundaries of inquiry are shaped by our willingness to recognize, embrace, and work with the unknown, the uncertain, and the inexpressible, transforming absence and incompleteness from limitations into productive forces**. This is especially critical in the current era of increasingly powerful AI: **only by acknowledging that AI cannot encompass all the potential wisdom that future humans may generate, can we escape a deterministic fate of being replaced by AI and continue to push the boundaries to create new forms of human intelligence that surpass AI**.